\documentclass[runningheads]{llncs}
\usepackage{graphicx}

\usepackage{tikz}
\usepackage{comment}
\usepackage{amsmath,amssymb} 
\usepackage{color}

\usepackage{microtype}
\usepackage{graphicx}
\usepackage{subfigure}
\usepackage{booktabs} 

\usepackage{nicefrac}       
\usepackage{microtype}      
\usepackage{array}
\usepackage[ruled, vlined]{algorithm2e}
\usepackage{bm}
\usepackage{multirow}
\usepackage{url}            
\usepackage{xcolor}
\usepackage{xspace}
\usepackage{arydshln}
\usepackage{caption}
\usepackage{wrapfig}
\usepackage{hyperref}

\usepackage[accsupp]{axessibility}  

\newcommand{\sysname}{DPU\xspace}

\newcommand{\eg}{\emph{e.g.},\xspace}
\newcommand{\ie}{\emph{i.e.},\xspace}

\newcommand\figref[1]{Fig.~\ref{#1}}

\newcommand\tabref[1]{Tab.~\ref{#1}}

\newcommand\secref[1]{Sec.~\ref{#1}}
\newcommand\equref[1]{Eq.(\ref{#1})}

\newcommand\imgtabcolsep{0.2em} 
\newcommand{\tabimga}[1]{\raisebox{-.5\height}{\includegraphics[width=0.29\textwidth]{{#1}.pdf}}}
\newcommand{\tabimgb}[1]{{\includegraphics[width=0.30\textwidth]{{#1}.pdf}}}
\newcommand{\tabimgc}[1]{{\includegraphics[width=0.30\textwidth,height=0.23\textwidth]{{#1}.pdf}}}
\newcommand{\fakeparagraph}[1]{\vspace{1mm}\noindent\textbf{#1.}}

\begin{document}
\pagestyle{headings}
\mainmatter
\def\ECCVSubNumber{1447}  

\title{Deep Partial Updating: Towards Communication Efficient Updating for On-device Inference} 

\titlerunning{Deep Partial Updating}
%
\author{Zhongnan Qu\inst{1} \and
Cong Liu\inst{2} \and
Lothar Thiele\inst{1}}
\authorrunning{Zhongnan Qu et al.}
%
\institute{ETH Zurich, Zurich, Switzerland\\
\email{\{quz,thiele\}@ethz.ch}\\
\and
UT Dallas, Texas, US\\
\email{cong@utdallas.edu}}
\maketitle

\begin{abstract}
Emerging edge intelligence applications require the server to retrain and update deep neural networks deployed on remote edge nodes to leverage newly collected data samples. Unfortunately, it may be impossible in practice to continuously send fully updated weights to these edge nodes due to the highly constrained communication resource. In this paper, we propose the weight-wise deep partial updating paradigm, which smartly selects a small subset of weights to update in each server-to-edge communication round, while achieving a similar performance compared to full updating. Our method is established through analytically upper-bounding the loss difference between partial updating and full updating, and only updates the weights which make the largest contributions to the upper bound. Extensive experimental results demonstrate the efficacy of our partial updating methodology which achieves a high inference accuracy while updating a rather small number of weights.
\end{abstract}

\section{Introduction}
\label{sec:introduction}

On-device inference is a new disruptive technology that enables new intelligent applications, \eg mobile assistants, Internet of Things and augmented reality. 
Compared to traditional cloud inference, on-device inference is subject to severe limitations in terms of storage, energy, computing power and communication. 
On the other hand, it has many advantages, \eg it enables fast and stable inference even with low communication bandwidth or interrupted communication, and can save energy by avoiding the transfer of data to the cloud, which often costs significant amounts of energy than sensing and computation \cite{bib:diss18:Varshney,bib:arXiv18:Guo,bib:arXiv19:Lee}.
To deploy deep neural networks (DNNs) on resource-constrained edge devices, extensive research has been done to compress a well pre-trained model via pruning \cite{bib:ICLR16:Han,bib:ICLR19:Frankle,bib:ICLR20:Renda} and quantization \cite{bib:NIPS15:Courbariaux,bib:ECCV16:Rastegari,bib:CVPR20:Qu}.
During on-device inference, compressed DNNs may achieve a good balance between model performance and resource demand.

However, due to a possible lack of relevant training data at the time of initial deployment or due to an unknown sensing environment, pre-trained DNNs may either fail to perform satisfactorily or be significantly improved after the initial deployment. 
In other words, re-training the models by using newly collected data (from \emph{edge devices} or \emph{other sources}) is typically required to achieve the desired performance during the lifetime of devices. 

Because of the resource-constrained nature of edge devices in terms of memory and computing power, on-device re-training (or federated learning) is typically restricted to tiny batch size, small inference (sub-)networks or limited optimization steps, all resulting in a performance degradation. 
Instead, retraining often occurs on a remote server with sufficient resources.
One possible strategy to allow for a continuous improvement of the model performance is a two-stage iterative process: (\textit{i}) at each round, edge devices collect new data samples and send them to the server, and (\textit{ii}) the server retrains the model using all collected data, and sends updates to each edge device \cite{bib:cs06:Brown}. 
The first stage may be even not necessary if new data are collected in other ways and made directly available to the server. 

\fakeparagraph{Example scenarios}
Example application scenarios of relevance include vision robotic sensing in an unknown environment (\eg Mars) \cite{bib:RAL17:Meng}, local translators of low-resource languages on mobile phones \cite{bib:ICML19:Bhandare,bib:arXiv20:Wang}, and sensor networks mounted in alpine areas \cite{bib:IPSN19:Meyer}, automatic wildlife monitoring \cite{bib:mee18:Stowell}.
We detail two specific scenarios.
\textit{Hazard alarming on mountains: }
Researchers in \cite{bib:IPSN19:Meyer} mounted tens of sensor nodes at different scarps in high alpine areas with cameras, geophones and high-precision GPS. The purpose is to achieve fast, stable, and energy-efficient hazard monitoring for early warning to protect people and infrastructure. 
To this end, a DNN is deployed on each node to on-device detect rockfalls and debris flows. 
The nodes regularly collect and send data to the server for labeling and retraining, and the server sends the updated model back through a low-power wireless network. Retraining during deployment is essential for a highly reliable hazard warning.  
\textit{Endangered species monitoring: }
To detect endangered species, researchers often deploy some audio or image sensor nodes in virgin rainforests \cite{bib:mee18:Stowell}.
Edge nodes are supposed to classify the potential signal from endangered species and send these relevant data to the server.
Due to the limited prior information from environments and species, retraining the initially classifier with received data or data from other sources (\eg other areas) is necessary.

\fakeparagraph{Challenges}
An essential challenge herein is that the transmissions in the server-to-edge stage are highly constrained by the limited communication resource (\eg bandwidth, energy \cite{bib:sensors16:Augustin}) in comparison to the edge-to-server stage, if necessary at all. 
Typically, state-of-the-art DNNs often require tens or even hundreds of mega-Bytes (MB) to store parameters, whereas a single batch of data samples (a number of samples that lead to reasonable updates in batch training) needs a relatively smaller amount of data. 
For example, for CIFAR10 dataset \cite{bib:cifar10}, the weights of a popular VGGNet require $56.09$MB storage, while one batch of 128 samples only uses around $0.40$MB \cite{bib:ICLR15:Simonyan,bib:ECCV16:Rastegari}. 
As an alternative approach, the server sends a full update of the inference model once or rarely. But in this case, every node will suffer from a low performance until such an update occurs.
Besides, edge devices could decide on and send only critical samples by using active learning schemes \cite{bib:ICLR20:Ash}.
The server may also receive data from other sources, \eg through data augmentation based on the data collected in previous rounds or new data collection campaigns.
These considerations indicate that the updated weights that are sent to edge devices by the server become a major bottleneck.

Facing the above challenges, we ask the following question: \textit{Is it possible to update only a small subset of weights while reaching a similar performance as updating all weights?}
Doing such a \textit{partial updating} can significantly reduce the server-to-edge communication overhead. 
Furthermore, fewer parameter updates also lead to less memory access on edge devices, which in turn results in smaller energy consumption than full updating \cite{bib:ISSCC14:Horowitz}. 

\fakeparagraph{Why partial updating works} 
Since the model deployed on edge devices is trained with the data collected beforehand, some learned knowledge can be reused. 
In other words, we only need to distinguish and update the weights which are critical to the newly collected data.

\fakeparagraph{How to select weights}
Our key concept for partial updating is based on the hypothesis, that \textit{a weight shall be updated only if it has a large contribution to the loss reduction} during the retraining given newly collected data samples.
Specially, we define a binary mask $\bm{m}$ to describe which weights are subject to update and which weights are fixed (also reused). 
For any $\bm{m}$, we establish the analytical upper bound on the difference between the loss value under partial updating and that under full updating.
We determine an optimized mask $\bm{m}$ by combining two different view points: (\textit{i}) measuring each weight's ``global contribution'' to the upper bound through computing the Euclidean distance, and (\textit{ii}) measuring each weight's ``local contribution'' to the upper bound using gradient-related information. 
The weights to be updated according to $\bm{m}$ will be further sparsely fine-tuned while the remaining weights are rewound to their initial values.

Our contributions can be summarized as follows.
\begin{itemize}
    \item We formalize the deep partial updating paradigm, \ie how to iteratively perform partial updating of inference models on remote edge devices, if newly collected training samples are available at the server. This reduces the computation and communication demand on edge devices substantially. 
    \item We propose a novel approach that determines the optimized subset of weights that shall be selected for partial updating, through measuring each weight's contribution to the analytical upper bound on the loss reduction. This simple yet effective metric can be applied to any models that are trained with gradient-based optimizers.
    \item Experimental results on public vision datasets show that, under the similar accuracy level along the rounds, our approach can reduce the size of the transmitted data by $95.3\%$ on average (up to $99.3\%$), namely can update the model averagely $21$ times more frequently than full updating.
\end{itemize}

\section{Related Work}
\label{sec:related}

\fakeparagraph{Partial updating}
Although partial updating has been adopted in some prior works, it is conducted in a fairly coarse-grained manner, \eg layer-wise or neuron-wise, and targets at completely different objectives. 
Especially, under continual learning settings, \cite{bib:ICLR18:Yoon,bib:arXiv20:Jung} propose to freeze all weights related to the neurons which are more critical in performing prior tasks than new ones, to preserve existing knowledge. 
Under adversarial attack settings, \cite{bib:CCS15:Shokri} updates the weights in the first several layers only, which yield a dominating impact on the extracted features, for better attack efficacy.
Under meta learning settings, \cite{bib:ICLR20:Raghu,bib:AAAI21:Shen} reuse learned representations by only updating a subset of layers for efficiently learning new tasks.
Unfortunately, such techniques do not focus on reducing the number of updated weights, and thus cannot be applied in our problem settings. 

\fakeparagraph{Federated learning}
Communication-efficient federated learning \cite{bib:ICLR18:Lin,bib:arXiv19:Kairouz,bib:arXiv20:Li} studies how to compress multiple gradients calculated on different sets of non-\textit{i.i.d.} local data, such that the aggregation of these compressed gradients could result in a similar convergence performance as centralized training on all data.
Such compressed updates are fundamentally different from our setting, where (\textit{i}) updates are not transmitted in each optimization step; (\textit{ii}) training data are incrementally collected; (\textit{iii}) centralized training is conducted.
Our typical scenarios focus on outdoor areas, which generally do not involve data privacy issues, since these collected data are not personal data. 
In comparison to federated learning, our pipeline has the following advantages: (\textit{i}) we do not conduct resource-intensive gradient backward propagation on edge devices; (\textit{ii}) the collected data are not continuously accumulated and stored on memory-constrained edge nodes; (\textit{iii}) we also avoid the difficult but necessary labeling process on each edge node in supervised learning tasks; (\textit{iv}) if few events occur on some nodes, the centralized training may avoid degraded updates in local training, \eg batch normalization.

\fakeparagraph{Compression}
The communication cost could also be reduced through some compression techniques, \eg quantizing/encoding the updated weights and the transmission signal \cite{bib:ICLR16:Han}. 
But note that these techniques are orthogonal to our approach and could be applied in addition, see Appendix \ref{app:pipeline}.

\fakeparagraph{Unstructured pruning}
Deep partial updating is inspired by recent unstructured pruning methods, \eg \cite{bib:ICLR16:Han,bib:ICLR19:Frankle,bib:NIPS19:Zhou,bib:ICLR20:Renda,bib:ICML21:Evci,bib:NIPS21:Peste}.
Traditional pruning methods aim at reducing the number of operations and storage consumption by setting some weights to zero. Sending a pruned DNN with only non-zero weights may also reduce the communication cost, but to a much lesser extent as shown in the experimental results, see \secref{sec:samplesRatio}. 
Since our objective namely reducing the server-to-edge communication cost when updating the deployed DNN is fundamentally different from pruning, we can leverage some learned knowledge by retaining weights (partial updating) instead of zero-outing weights (pruning). 

\fakeparagraph{Domain adaptation}
Domain adaptation targets reducing domain shift to transfer knowledge into new learning tasks \cite{bib:arXiv19:Zhuang}.
This paper mainly considers the scenario where the inference task is not explicitly changed along the rounds, \ie the overall data distribution maintains the same along the data collection rounds. 
Thus, selecting critical weights (features) by measuring their impact on domain distribution discrepancy is invalid herein. 
Applying deep partial updating on streaming tasks where the data distribution varies along the rounds would be also worth studying, and we leave it for future works.

\section{Notation and Setting}
\label{sec:notation}

In this section, we define the notations used throughout this paper, and provide a formalized problem setting. 
We consider a set of remote edge devices that implement on-device inference. 
They are connected to a host server that is able to perform DNN training and retraining. 
We consider the necessary amount of information that needs to be communicated to each edge device to update its inference model. 

Assume there are $R$ rounds of model updates. 
The model deployed in the $r$-th round is represented with its weight vector $\bm{w}^r$. 
The training data used to update the model for the $r$-th round is represented as $\mathcal{D}^r = \delta\mathcal{D}^{r}\cup\mathcal{D}^{r-1}$. 
Also, newly collected data samples $\delta\mathcal{D}^r$ are made available to the server in round $r-1$. 

To reduce the amount of information that needs to be sent to edge devices, only partial weights of $\bm{w}^{r-1}$ shall be updated when determining $\bm{w}^{r}$. 
The overall optimization problem for weight-wise partial updating in round $r-1$ is thus,
\begin{eqnarray}
    \min_{\delta\bm{w}^r}   & & \ell\left(\bm{w}^{r-1}+\delta\bm{w}^{r};\mathcal{D}^r\right) \label{eq:objective_r} \\
    \text{s.t.}             & & \|\delta\bm{w}^{r}\|_0 \leq k \cdot I \label{eq:constraints_r}
\end{eqnarray}
where $\ell$ denotes the loss function, $\|.\|_0$ denotes the L0-norm, $k$ denotes the updating ratio that is determined by the communication constraints in practical scenarios, and $\delta\bm{w}^{r}$ denotes the increment of $\bm{w}^{r-1}$. 
Note that both $\bm{w}^{r-1}$ and $\delta\bm{w}^{r}$ are drawn from $\mathbb{R}^I$, where $I$ is the total number of weights. 

In this case, only a fraction of $k \cdot I$ weights and the corresponding index information need to be communicated to each edge device for updating the model, namely the partial updates $\delta\bm{w}^{r}$. 
It is worth noting that the index information is relatively small in size compared to the partially updated weights (see \secref{sec:experiment}).
On each edge device, the weight vector is updated as $\bm{w}^{r} = \bm{w}^{r-1}+\delta\bm{w}^{r}$.
To simplify the notation, we will only consider a single update, \ie from weight vector $\bm{w}$ (corresponding to $\bm{w}^{r-1}$) to weight vector $\widetilde{\bm{w}}$ (corresponding to $\bm{w}^{r}$) with $\widetilde{\bm{w}} = \bm{w}+\widetilde{\delta\bm{w}}$.  

\section{Deep Partial Updating}
\label{sec:partialUpdating}

\begin{wrapfigure}{r}{0.5\textwidth}
	\includegraphics[width=0.49\textwidth]{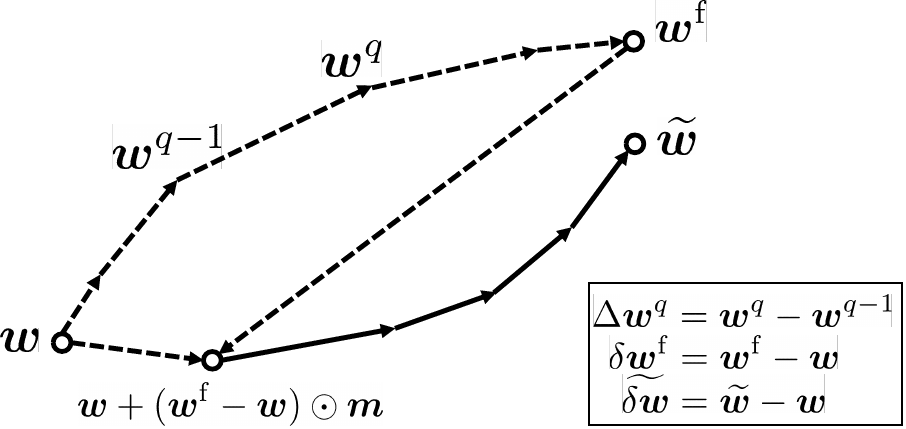}
	\caption{The figure depicts the overall approach that consists of two steps. The first step is depicted with dotted arrows and starts from the deployed model $\bm{w}$. In $Q$ optimization steps, all weights are trained to the optimum $\bm{w}^\mathrm{f}$. Based on the collected information, a binary mask $\bm{m}$ is determined that characterizes the set of weights that are rewound to the ones of $\bm{w}$. Therefore, the second step (solid arrows) starts from $\bm{w} + \delta\bm{w}^\mathrm{f} \odot \bm{m}$. According to the mask, this solution is sparsely fine-tuned to the final weights $\widetilde{\bm{w}}$, \ie $\widetilde{\delta\bm{w}}$ has only non-zero values where the mask value is 1}
	\label{fig:approach}
\end{wrapfigure}

We developed a two-step approach for resolving the partial updating optimization problem in \equref{eq:objective_r}-\equref{eq:constraints_r}. 
The overall approach is depicted in \figref{fig:approach}.

\fakeparagraph{The first step} The first step not only determines the subset of weights that are allowed to change their values, but also computes the initial values for the second step. 
In particular, we first optimize the loss function $\ell$ by updating all weights from the initialization $\bm{w}$ with a standard optimizer, \eg SGD or its variants. 
We thus obtain the minimized loss $\ell\left(\bm{w}^\mathrm{f}\right)$ with $\bm{w}^\mathrm{f} = \bm{w} + \delta\bm{w}^\mathrm{f}$, where the superscript $\mathrm{f}$ denotes ``full updating''. 
To consider the constraint of \equref{eq:constraints_r}, the information gathered during this optimization is used to determine the subset of weights that will be changed, also that are communicated to the edge devices. 
In the explanation of the method in \secref{sec:metric}, we use the binary mask $\bm{m}$ with $\bm{m} \in\{0,1\}^I$ to describe which weights are subject to change and which ones are not. 
The weights with $m_i=1$ are trainable, whereas the weights with $m_i=0$ will be rewound from the values in $\bm{w}^\mathrm{f}$ to their initial values in $\bm{w}$, \ie unchanged. 
Obviously, we find $\|\bm{m}\|_0 = \sum_{i} m_i = k \cdot I$. 

\fakeparagraph{The second step} In the second step we start a sparse fine-tuning from a DNN with $k \cdot I$ weights from the optimized model $\bm{w}^\mathrm{f}$ and $(1-k) \cdot I$ weights from the previous, still deployed model $\bm{w}$. 
In other words, the initial weights for the second step are $\bm{w} + \delta\bm{w}^{\mathrm{f}} \odot \bm{m}$, where $\odot$ denotes an element-wise multiplication. 
To determine the final solution $\widetilde{\bm{w}} = \bm{w}+\widetilde{\delta\bm{w}}$, we conduct a sparse fine-tuning (still with a standard optimizer), \ie we keep all weights with $m_i = 0$ constant during the optimization. 
Therefore, $\widetilde{\delta\bm{w}}$ is zero wherever $m_i=0$, and only weights where $m_i = 1$ are updated.

\subsection{Metrics for Rewinding}
\label{sec:metric}

We will now describe a new metric that determines the weights that should be kept constant, \ie with $m_i=0$. 
Like most learning methods, we focus on minimizing a loss function.
The two-step approach relies on the following assumption: the better the loss $\ell(\bm{w} + \delta\bm{w}^\mathrm{f} \odot \bm{m})$ of the initial solution for the second step, the better the final performance.
Therefore, the first step should select a mask $\bm{m}$ such that the loss difference $\ell(\bm{w} + \delta\bm{w}^\mathrm{f} \odot \bm{m}) - \ell(\bm{w}^\mathrm{f})$ is as small as possible. 

To determine an optimized mask $\bm{m}$, we propose to upper-bound the above loss difference in two view points, and measure each weight's contribution to the bounds. 
The ``global contribution'' uses the norm information of incremental weights $\delta\bm{w}^\mathrm{f}=\bm{w}^\mathrm{f}-\bm{w}$.
The ``local contribution'' takes into account the gradient-based information that is gathered during the optimization in the first step, \ie in the path from $\bm{w}$ to $\bm{w}^\mathrm{f}$. 
Both contributions will be combined to determine the mask $\bm{m}$.

The two view points are based on the concept of smooth differentiable functions. 
A function $f(x)$ with $f: \mathbb{R}^d \rightarrow \mathbb{R}$ is called $L$-smooth if it has a Lipschitz continuous gradient $g(x)$: $\|g(x) - g(y)\|_2 \leq L \|x - y\|_2$ for all $x, y$. 
Note that Lipschitz continuity of gradients is essential to ensuring convergence of many gradient-based algorithms. 
Under such a condition, one can derive the following bounds, see also \cite{bib:book98:Nesterov}:
\begin{equation} \label{eq:lipschitz}
    |f(y) - f(x) - g(x)^\mathrm{T} \cdot (y - x) | \leq L/2 \cdot \|y - x\|_2^2 \quad \forall x, y
\end{equation}

\fakeparagraph{Global contribution}
One would argue that a large absolute value in $\delta\bm{w}^\mathrm{f} = \bm{w}^\mathrm{f} - \bm{w}$ indicates that this weight has moved far from its initial value in $\bm{w}$, and thus should not be rewound.
This motivates us to adopt the widely used unstructured magnitude pruning to determine the mask $\bm{m}$.
Magnitude pruning prunes the weights with the lowest magnitudes, which often achieves a good trade-off between the model accuracy and the number of zero's weights \cite{bib:ICLR20:Renda}. 

Using $a - b \leq |a - b|$, \equref{eq:lipschitz} can be reformulated as $f(y) - f(x) - g(x)^T (y-x) \leq | f(y) - f(x) - g(x)^T (y-x) | \leq L/2 \cdot \| y-x \|^2_2$.
Thus, we can bound the relevant loss difference $\ell(\bm{w} + \delta\bm{w}^\mathrm{f} \odot \bm{m}) - \ell(\bm{w}^\mathrm{f}) \geq 0$ as
\begin{equation} \label{eq:globalbound}
    \ell(\bm{w} + \delta\bm{w}^\mathrm{f} \odot \bm{m}) - \ell(\bm{w}^\mathrm{f}) \leq \bm{g}(\bm{w}^\mathrm{f})^\mathrm{T} \cdot \left( \delta\bm{w}^\mathrm{f} \odot (\bm{m} - \bm{1}) \right) + L/2 \cdot \| \delta\bm{w}^\mathrm{f} \odot (\bm{m} - \bm{1})\|_2^2
\end{equation}
where $\bm{g}(\bm{w}^\mathrm{f})$ denotes the loss gradient at $\bm{w}^\mathrm{f}$. 
As the loss is optimized at $\bm{w}^\mathrm{f}$, \ie $\bm{g}(\bm{w}^\mathrm{f})\approx \bm{0}$, we can assume that the gradient term is much smaller than the norm of the weight differences in \equref{eq:globalbound}. 
Therefore, we have
\begin{equation} \label{eq:globalsum}
    \ell(\bm{w} + \delta\bm{w}^\mathrm{f} \odot \bm{m}) - \ell(\bm{w}^\mathrm{f}) \lesssim L/2 \cdot \| \delta\bm{w}^\mathrm{f} \odot (\bm{1} - \bm{m})\|_2^2
\end{equation}
The right hand side is clearly minimized if $m_i = 1$ for the largest absolute values of $\delta\bm{w}^\mathrm{f}$. 
As $\bm{1}^\mathrm{T} \cdot \left( \bm{c}^\mathrm{global} \odot (\bm{1} - \bm{m}) \right) = \| \delta\bm{w}^\mathrm{f} \odot (\bm{1} - \bm{m})\|_2^2$, this information is captured in the contribution vector 
\begin{equation} \label{eq:globalc}
    \bm{c}^\mathrm{global} = \delta\bm{w}^\mathrm{f} \odot \delta\bm{w}^\mathrm{f}
\end{equation}
The $k \cdot I$ weights with the largest values in $\bm{c}^\mathrm{global}$ are assigned to mask values $1$ and are further fine-tuned in the second step, whereas all others are rewound to their initial values in $\bm{w}$. 
Alg. \ref{alg:gcpu} in Appendix \ref{app:gcpu} shows this first approach.

\fakeparagraph{Local contribution}
As experiments show, one can do better when leveraging in addition some gradient-based information gathered during the first step, \ie optimizing the initial weights $\bm{w}$ in $Q$ traditional optimization steps, $\bm{w} = \bm{w}^0 \rightarrow \cdots \; \rightarrow \bm{w}^{q-1} \rightarrow \bm{w}^{q} \rightarrow \cdots \; \rightarrow \bm{w}^{Q} = \bm{w}^\mathrm{f}$.

Using $ -a + b \leq | a - b|$, \equref{eq:lipschitz} can be reformulated as $f(x) - f(y) + g(x)^T (y-x) \leq | f(y) - f(x) - g(x)^T (y-x) | \leq L/2 \cdot \| y-x \|^2_2$. 
Thus, each optimization step is bounded as
\begin{equation} \label{eq:localbound}
    \ell(\bm{w}^{q-1}) - \ell(\bm{w}^q) \leq -\bm{g}(\bm{w}^{q-1})^\mathrm{T} \cdot \Delta\bm{w}^q + L/2 \cdot \| \Delta\bm{w}^q \|_2^2
\end{equation}
where $\Delta\bm{w}^q = \bm{w}^q - \bm{w}^{q-1}$. 
For a conventional gradient descent optimizer with a small learning rate we can use the approximation $|\bm{g}(\bm{w}^{q-1})^\mathrm{T} \cdot \Delta\bm{w}^q| \gg \|\Delta\bm{w}^q\|_2^2$ and obtain $ \ell(\bm{w}^{q-1}) - \ell(\bm{w}^q) \lesssim -\bm{g}(\bm{w}^{q-1})^\mathrm{T} \cdot \Delta\bm{w}^q$. 
Summing up over all optimization iterations yields approximately
\begin{equation} \label{eq:localsum}
    \ell(\bm{w}^\mathrm{f} - \delta\bm{w}^\mathrm{f}) - \ell(\bm{w}^\mathrm{f}) \lesssim -\sum_{q = 1}^Q \bm{g}(\bm{w}^{q-1})^\mathrm{T} \cdot \Delta\bm{w}^q
\end{equation}
Note that we have $\bm{w} = \bm{w}^\mathrm{f} - \delta\bm{w}^\mathrm{f}$ and $\delta\bm{w}^\mathrm{f} = \sum_{q = 1}^Q \Delta\bm{w}^q$. 
Therefore, with a small updating ratio $k$, \ie $\bm{m} \sim \bm{0}$, we can reformulate \equref{eq:localsum} as
$ 
    \ell\left( \bm{w} + \delta\bm{w}^\mathrm{f} \odot \bm{m} \right) - \ell(\bm{w}^\mathrm{f}) \lesssim \mathrm{U}(\bm{m}) 
$
with the upper bound 
$
    \mathrm{U}(\bm{m}) = -\sum_{q = 1}^Q \bm{g}(\bm{w}^{q-1})^\mathrm{T} \cdot (\Delta\bm{w}^q \odot (\bm{1} - \bm{m}))
$
where we suppose that the gradients are approximately constant for $\bm{m} \sim \bm{0}$ (\ie $\bm{m}$ has zero entries almost everywhere).
Therefore, an approximate incremental contribution of each weight dimension to the upper bound on the loss difference $\ell\left( \bm{w} + \delta\bm{w}^\mathrm{f} \odot \bm{m} \right) - \ell(\bm{w}^\mathrm{f})$ 
can be determined by the negative gradient vector at $\bm{m}=\bm{0}$, denoted as 
\begin{equation}
    \bm{c}^{\mathrm{local}} = - \frac{\partial \mathrm{U}(\bm{m})}{\partial \bm{m}} = - \sum_{q=1}^{Q} \bm{g}(\bm{w}^{q-1}) \odot \Delta\bm{w}^{q}
\end{equation}
which models the accumulated contribution to the overall loss reduction.

\fakeparagraph{Combining global and local contribution}
So far, we independently calculate the global and local contributions. 
To avoid the scale impact, we first normalize each contribution by its significance in its own set (either global or local contribution set). 
We investigated the impacts and the different combinations of both normalized contributions, see the results in Appendix \ref{app:impact}.
Interestingly, the most straightforward combination (\ie the sum of both normalized metrics) often yields a satisfied and stable performance.
Intuitively, local contribution can better identify critical weights w.r.t. the loss during training, while global contribution may be more robust for a highly non-convex loss landscape. 
Both metrics may be necessary when selecting weights to rewind.
Therefore, the combined contribution is computed as
\begin{equation}
    \bm{c} = \frac{1}{\bm{1}^\mathrm{T} \cdot \bm{c}^\mathrm{global}} \bm{c}^\mathrm{global} + \frac{1}{\bm{1}^\mathrm{T} \cdot \bm{c}^\mathrm{local}} \bm{c}^\mathrm{local}
    \label{eq:combined}
\end{equation}
and $m_i = 1$ for the $k \cdot I$ largest values of $\bm{c}$ and $m_i = 0$ otherwise. 
The pseudocode of Deep Partial Updating (\sysname), \ie rewinding according to the combined contribution to the loss reduction, is shown in Alg. \ref{alg:dpu}.
The complexity analysis of the corresponding algorithm are shown in Appendix \ref{app:complexity}.

\begin{algorithm}[tbp!]
    \caption{Deep Partial Updating}\label{alg:dpu}
    \KwIn{weights $\bm{w}$, updating ratio $k$, learning rate $\{\alpha^q\}_{q=1}^{Q}$ in $Q$ iterations} 
    \KwOut{weights $\widetilde{\bm{w}}$}
    \tcc{The first step: full updating and rewinding}
    Initiate $\bm{w}^0=\bm{w}$ and $\bm{c}^{\mathrm{local}}=\bm{0}$\;
    \For {$q \leftarrow 1$ \KwTo $Q$} {
        Compute the loss gradient $\bm{g}(\bm{w}^{q-1})=\partial\ell(\bm{w}^{q-1})/\partial\bm{w}^{q-1}$\;
        Compute the optimization step with learning rate $\alpha^q$ as $\Delta\bm{w}^{q}$\;
        Update $\bm{w}^{q} = \bm{w}^{q-1} + \Delta\bm{w}^{q}$\;
        Update $\bm{c}^{\mathrm{local}}=\bm{c}^{\mathrm{local}}-\bm{g}(\bm{w}^{q-1})\odot\Delta\bm{w}^{q}$\;
    }
    Set $\bm{w}^{\mathrm{f}}=\bm{w}^{Q}$ and get $\delta\bm{w}^{\mathrm{f}} = \bm{w}^{\mathrm{f}}-\bm{w}$\;
    Compute $\bm{c}^{\mathrm{global}}=\delta\bm{w}^{\mathrm{f}}\odot\delta\bm{w}^{\mathrm{f}}$\;
    Compute $\bm{c}$ as \equref{eq:combined} and sort in descending order\;
    Create a binary mask $\bm{m}$ with $1$ for the Top-$(k \cdot I)$ indices, $0$ for others\;
    \tcc{The second step: sparse fine-tuning}
    Initiate $\widetilde{\delta\bm{w}}=\delta\bm{w}^{\mathrm{f}}\odot\bm{m}$ and $\widetilde{\bm{w}} = \bm{w}+\widetilde{\delta\bm{w}}$\;
    \For {$q \leftarrow 1$ \KwTo $Q$} {
        Compute the optimization step on $\widetilde{\bm{w}}$ with learning rate $\alpha^q$ as $\Delta\widetilde{\bm{w}}^{q}$\; 
        Update $\widetilde{\delta\bm{w}}=\widetilde{\delta\bm{w}}+\Delta\widetilde{\bm{w}}^{q}\odot\bm{m}$ and $\widetilde{\bm{w}}=\bm{w}+\widetilde{\delta\bm{w}}$\;
    }
\end{algorithm}

\subsection{(Re-)Initialization of Weights}
\label{sec:initialization}

In this section, we discuss the initialization of \sysname.
$\mathcal{D}^1$ denotes the initial dataset used to train the model $\bm{w}^1$ from a randomly initialized model $\bm{w}^0$. 
$\mathcal{D}^1$ corresponds to the available dataset before deployment, or collected in the $0$-th round if there are no data available before deployment. 
$\{\delta\mathcal{D}^r\}_{r=2}^R$ denotes newly collected samples in each subsequent round. 

Experimental results show (see Appendix \ref{app:fullupdating}) that training from a randomly initialized model can yield a higher accuracy \textit{after a large number of rounds}, compared to always training from the last round $\bm{w}^{r-1}$.
As a possible explanation, the optimizer could end in a hard to escape region of the search space if always trained from the last round for a long sequence of rounds.
Thus, we propose to re-initialize the weights after a certain number of rounds. 
In such a case, Alg. \ref{alg:dpu} does not start from the weights $\bm{w}^{r-1}$ but from the randomly initialized weights.
The randomly re-initialized model (weights) can be efficiently sent to the edge devices via a single random seed. 
The device can determine the weights by means of a random generator. 
This process realizes a random shift in the search space, which is a communication-efficient way in comparison to other alternatives, such as learning to increase the loss or using the (averaged) weights in the previous rounds, as these fully changed weights still need to be sent to each node.
Each time the model is randomly re-initialized, the new partially updated model might suffer from an accuracy drop in a few rounds. 
However, we can simply avoid such an accuracy drop by not updating the model if the validation accuracy does not increase compared to the last round, see Appendix \ref{app:reinit}.
Note that the learned knowledge thrown away by re-initialization can be re-learned afterwards, since all collected samples are continuously stored and accumulated in the server. 
This also makes our setting different from continual learning, that aims at avoiding catastrophic forgetting without accessing old data.

To determine after how many rounds the model should be re-initialized, we conduct extensive experiments on different partial updating settings, see Appendix \ref{app:reinit}.
In conclusion, the model is randomly re-initialized as long as the number of total newly collected data samples exceeds the number of samples when the model was re-initialized last time.
For example, assume that at round $r$ the model is randomly (re-)initialized and partially updated from this random model on dataset $\mathcal{D}^r$. 
Then, the model will be re-initialized again at round $r+n$, if $|\mathcal{D}^{r+n}|>2\cdot|\mathcal{D}^r|$, where $|.|$ denotes the number of samples in the dataset.

\section{Evaluation}
\label{sec:experiment}

In this section, we experimentally show that through updating a small subset of weights, \sysname can reach a similar accuracy as full updating while requiring a significantly lower communication cost. 
We implement \sysname with Pytorch \cite{bib:NIPSWorkshop17:Paszke}, and evaluate on public vision datasets, including MNIST \cite{bib:MNIST}, CIFAR10 \cite{bib:cifar10}, CIFAR100 \cite{bib:cifar10}, ImageNet \cite{bib:ILSVRC15}, using multilayer perceptron (MLP), VGGNet \cite{bib:NIPS15:Courbariaux,bib:ECCV16:Rastegari,bib:CVPR20:Qu}, ResNet56 \cite{bib:CVPR16:He}, MobileNetV1 \cite{bib:arXiv17:Howard}, respectively. 
Particularly, we partition the experiments into multi-round updating and single-round updating.

\fakeparagraph{Multi-round updating} 
We consider there are limited (or even zero) samples before the initial deployment, and data samples are continuously collected and sent from edge devices over a long period (the event rate is often low in real cases \cite{bib:IPSN19:Meyer}). 
The server retrains the model and sends the updates to each device in multiple rounds.
Regarding the highly-constrained communication resources, we choose low resolution image datasets (MNIST \cite{bib:MNIST} and CIFAR10/100 \cite{bib:cifar10}) to evaluate multi-round updating. 
We conduct one-shot rewinding in multi-round \sysname, \ie rewinding is executed only once to achieve the desired updating ratio at each round as in Alg. \ref{alg:dpu}, which avoids hand-tuning hyperparameters (\eg updating ratio schedule) frequently over a large number of rounds.

\fakeparagraph{Single-round updating} 
The deployed model is updated once via server-to-edge communication when new data from other sources become available on the server after some time, \eg releasing a new version of mobile applications based on newly retrieved internet data.
Although \sysname is elaborated and designed under multi-round updating settings, it can be applied directly in single-round updating. 
Since transmission from edge devices may be even not necessary, we evaluate single-round \sysname on the large scale ImageNet dataset. 
Iterative rewinding is adopted here due to its better performance. 
Particularly, we alternatively perform rewinding 20\% of the remaining trainable weights according to \equref{eq:combined} and sparse fine-tuning until reaching the desired updating ratio. 

\fakeparagraph{Settings for all experiments} 
We randomly select 30\% of the original test dataset (original validation dataset for ImageNet) as the validation dataset, and the remainder as the test dataset.
Let $\{|\mathcal{D}^1|,|\delta\mathcal{D}^r|\}$ represent the available data samples along rounds, where $|\delta\mathcal{D}^r|$ is supposed to be constant along rounds.
Both $\mathcal{D}^1$ and $\delta\mathcal{D}^r$ are randomly sampled (without replacement) from the original training dataset to simulate the data collection.
In each round, the test accuracy is reported, when the validation dataset achieves the highest Top-1 accuracy during retraining.
When the validation accuracy does not increase compared to the previous round, the models are not updated to reduce the communication overhead.
This strategy is also applied to other baselines to enable a fair comparison.  
We use the average cross-entropy as the loss function, and use Adam variant of SGD for MLP and VGGNet, Nesterov SGD for ResNet56 and MobileNetV1.
More implementation details are provided in Appendix \ref{app:implementation}.

\begin{figure}[t]
    \setlength\tabcolsep{\imgtabcolsep}
    \centering
    \begin{tabular}{ccc}
        \tabimgc{figure/mlp-pruning}    & \tabimgc{figure/vgg-pruning}      & \tabimgc{figure/resnet56-pruning}      \\ 
    \end{tabular}
    \caption{\sysname is compared with other baselines on different benchmarks in terms of the test accuracy during multi-round updating}
    \label{fig:multiround}
\end{figure}

\fakeparagraph{Indexing}
\sysname generates a sparse tensor.
In addition to the updated weights, the indices of these weights also need to be sent to each edge device.
A simple implementation is to send the mask $\bm{m}$, \ie a binary vector of $I$ elements.
Let $S_w$ denote the bitwidth of each single weight, and $S_x$ denote the bitwidth of each index.
Directly sending $\bm{m}$ yields an overall communication cost of $I\cdot k \cdot S_w+I \cdot S_x$ with $S_x=1$. 
To save the communication cost on indexing, we further encode $\bm{m}$. Suppose that $\bm{m}$ is a random binary vector with a probability of $k$ to contain 1. 
The optimal encoding scheme according to Shannon yields $S_x(k)=k \cdot \mathrm{log}(1/k) + (1-k) \cdot \mathrm{log}(1/(1-k))$. 
Coding schemes such as Huffman block coding can come close to this bound. 
We use $S_w\cdot k\cdot I + S_x(k)\cdot I$ to report the size of data transmitted from server to each node at each round, contributed by the partially updated weights plus the encoded indices of these weights.

\subsection{Benchmarking Multi-Round Updating}
\label{sec:multiround}

\fakeparagraph{Settings} 
To the best of our knowledge, this is the first work on studying weight-wise partial updating a model using newly collected data in iterative rounds. 
Therefore, we developed three baselines for comparison, including (\textit{i}) full updating (FU), where at each round the model is fully updated from a random initialization (\ie training from scratch, which yields a better performance see \secref{sec:initialization} and Appendix \ref{app:fullupdating}); (\textit{ii}) random partial updating (RPU), where the model is trained from $\bm{w}^{r-1}$, while we randomly fix each layer's weights with a ratio of $(1-k)$ and sparsely fine-tune the rest; (\textit{iii}) global contribution partial updating (GCPU), where the model is trained with Alg. \ref{alg:gcpu} without re-initialization described in \secref{sec:initialization}; (\textit{iv}) a state-of-the-art unstructured pruning method \cite{bib:ICLR20:Renda}, where the model is first trained from a random initialization at each round, then conducts one-shot magnitude pruning, and finally is sparsely fine-tuned with learning rate rewinding. 
The ratio of nonzero weights in pruning is set to the same as the updating ratio $k$ to ensure the same communication cost.
The experiments are conducted on different benchmarks as mentioned earlier.

\begin{table}[t]
    \centering
    \caption{The average accuracy difference over all rounds and the ratio of communication cost over all rounds related to full updating} 
    \label{tab:multiround}
    \begin{tabular}{ccccccc}
    \toprule
    \multirow{2}{* }{Method}          & \multicolumn{3}{c}{Average accuracy difference}           & \multicolumn{3}{c}{Ratio of communication cost}       \\ \cmidrule(lr){2-4}  \cmidrule(lr){5-7} 
                                      & MLP               & VGGNet            & ResNet56          & MLP           & VGGNet        & ResNet56              \\ \hline
    \sysname                          & $\bm{-0.17\%}$    & $\bm{+0.33\%}$    & $\bm{-0.42\%}$    & $0.0071$     & $0.0183$       & $0.1147$              \\
    GCPU                              & $-0.72\%$         & $-1.51\%$         & $-3.87\%$         & $0.0058$     & $0.0198$       & $0.1274$              \\
    RPU                               & $-4.04\%$         & $-11.35\%$        & $-7.78\%$         & $0.0096$     & $0.0167$       & $0.1274$              \\
    Pruning \cite{bib:ICLR20:Renda}   & $-1.45\%$         & $-4.35\%$         & $-2.35\%$         & $0.0106$     & $0.0141$       & $0.1274$              \\
    \bottomrule
    \end{tabular}
\end{table}

\fakeparagraph{Results}
We report the test accuracy of the model $\bm{w}^r$ along rounds in \figref{fig:multiround}.
All methods start from the same $\bm{w}^0$, an entirely randomly initialized model.
As seen in this figure, \sysname clearly yields the highest accuracy in comparison to the other partial updating schemes.
For example, \sysname can yield a final Top-1 accuracy of $92.85\%$ on VGGNet, even exceeds the accuracy ($92.73\%$) of full updating. 
In addition, we compare three partial updating schemes in terms of the accuracy difference related to full updating averaged over all rounds, and the ratio of the communication cost related to full updating over all rounds in \tabref{tab:multiround}. 
As seen in the table, \sysname reaches a similar accuracy as full updating, while incurring significantly fewer transmitted data sent from the server to each edge node.
Specially, \sysname saves around $99.3\%$, $98.2\%$ and $88.5\%$ of transmitted data on MLP, VGGNet, and ResNet56, respectively ($95.3\%$ in average). 
The communication cost ratios shown in \tabref{tab:multiround} differ a little even for the same updating ratio. 
This is because if the validation accuracy does not increase, the model will not be updated to reduce the communication cost, as discussed earlier.

We further investigate the benefit due to \sysname in terms of \textit{the total communication cost reduction}, as \sysname has no impact on the edge-to-server communication involving sending newly collected data samples.
This experimental setup assumes that all samples in $\delta\mathcal{D}^r$ are collected by $N$ edge nodes during all rounds and sent to the server on a per-round basis.
For clarity, let $S_d$ denote the data size of each training sample.
During round $r$, we define per-node communication cost under \sysname as $S_d\cdot|\delta\mathcal{D}^r|/N+(S_w\cdot k\cdot I+S_x(k)\cdot I)$. 
The results are shown in Appendix \ref{app:totalcost}.
We observe that \sysname can still achieve a significant reduction on the total communication cost, \eg reducing up to $88.2\%$ even for the worst case (a single node). 
Moreover, \sysname tends to be more beneficial when the size of data transmitted by each node to the server becomes smaller. 
This is intuitive because in this case server-to-edge communication cost (thus the reduction due to \sysname) dominates the total communication cost. 

\begin{figure}[t]
    \setlength\tabcolsep{\imgtabcolsep}
    \centering
    \begin{tabular}{m{0.3cm}ccc}
                                      & \textbf{~~~~~~\{1000,~5000\}}         & \textbf{~~~~~~\{5000,~1000\}}         & \textbf{~~~~~~\{1000,~1000\}}             \\ 
    \rotatebox{90}{\textbf{0.1}}      & \tabimga{figure/1000-5000-3-pruningd} & \tabimga{figure/5000-1000-3-pruningd} & \tabimga{figure/1000-1000-3-pruningd}     \\ 
    \rotatebox{90}{\textbf{0.01}}     & \tabimga{figure/1000-5000-1-pruningd} & \tabimga{figure/5000-1000-1-pruningd} & \tabimga{figure/1000-1000-1-pruningd}     \\ 
    \end{tabular}
    \caption{Comparison w.r.t. the mean accuracy difference (full updating as the reference) under different $\{|\mathcal{D}^1|,|\delta\mathcal{D}^r|\}$ (representing the available data samples along rounds, see in \secref{sec:experiment}) and updating ratio ($k=0.1,0.01$)}
    \label{fig:number_ratio}
\end{figure}

\subsection{Different Number of Data Samples and Updating Ratios}
\label{sec:samplesRatio}

\fakeparagraph{Settings}
In this section, we show that \sysname outperforms other baselines under varying number of training samples and updating ratios in multi-round updating.
We also conduct ablations concerning the re-initialization of weights discussed in \secref{sec:initialization}.
We implement \sysname with and without re-initialization, GCPU with and without re-initialization, RPU, and Pruning \cite{bib:ICLR20:Renda} (see more details in \secref{sec:multiround}) on VGGNet using CIFAR10 dataset. 
We compare these methods with different amounts of samples $\{|\mathcal{D}^1|,|\delta\mathcal{D}^r|\}$ and different updating ratios $k$.
Without further notations, each experiment runs three times using random data samples.

\fakeparagraph{Results}
We compare the difference between the accuracy under each method and that under full updating. The mean accuracy difference over three runs is plotted in \figref{fig:number_ratio}.
A comprehensive set of results including the absolute accuracy and the standard deviations is provided in Appendix \ref{app:ablation}.
As seen in \figref{fig:number_ratio}, \sysname (with re-initialization) always achieves the highest accuracy. 
\sysname also significantly outperforms the pruning method, especially under a small updating ratio. 
Note that we preferred a smaller updating ratio in our context because it explores the limits of the approach and it indicates that we can improve the deployed model more frequently with the same accumulated server-to-edge communication cost.
The dashed curves and the solid curves with the same color can be viewed as the ablation study of our re-initialization scheme. 
Particularly given a large number of rounds, it is critical to re-initialize the start point $\bm{w}^{r-1}$ after several rounds (as discussed in \secref{sec:initialization}). 

In the first few rounds, partial updating methods almost always yield a higher test accuracy than full updating, \ie the curves are above zero. 
This is due to the fact that the amount of available samples is rather small, and partial updating may avoid some co-adaptation in full updating.
The partial updating methods perform almost randomly in the first round compared to each other, because the limited data are not sufficient to distinguish critical weights from the random initialization $\bm{w}^0$.
This also motivates us to (partially) update the deployed model when new samples are available.

\fakeparagraph{Pruning weights vs. pruning incremental weights}
One of our chosen baselines, global contribution partial updating (GCPU, see Alg. \ref{alg:gcpu}), could be viewed as a counterpart of the pruning method \cite{bib:ICLR20:Renda}, \ie pruning the incremental weights with the least magnitudes.
By comparing GCPU (with or without re-initialization) with ``Pruning'', we conclude that retaining previous weights yields better performance than zero-outing the weights.

\subsection{Benchmarking Single-Round Updating}
\label{sec:singleround}

\fakeparagraph{Settings}
To show the versatility of our methods, we test single-round updating for MobileNetV1 \cite{bib:arXiv17:Howard} on ImageNet \cite{bib:ILSVRC15} with iterative rewinding. 
Single-round \sysname is conducted on different initially deployed models, including a floating-point (FP32) dense model and two compressed models, \ie a 50\%-sparse model and an INT8 quantized model.
The sparse model is trained with a state-of-the-art dynamic pruning method \cite{bib:NIPS21:Peste}; the quantized model is trained with straight-through-estimator with a output-channel-wise floating point scaling factors similar as \cite{bib:ECCV16:Rastegari}. 
To maintain the same on-device inference cost, partial updating is only applied on nonzero values of sparse models; for quantized models, the updated weights are still in INT8 format. 

\begin{wraptable}{r}{0.62\textwidth}
    \caption{The test accuracy of single-round updating on different initial models (MobileNetV1 on ImageNet). The updating ratio $k=0.2$. The ratio of communication cost related to full updating is reported in brackets} 
    \label{tab:singleround}
    \centering
    \begin{tabular}{cccccc}
    \toprule
    \#Samples       & \multicolumn{3}{c}{$\{8\times10^5,4.8\times10^5\}$}   \\ 
                    \cmidrule(lr){2-4}  
                    & Initial   & Vanilla-update    & \sysname              \\ \hline      
    FP32 Dense      & 68.5\%    & 70.7\% (1)        &  71.1\% (0.22)        \\ 
    50\%-Sparse     & 68.1\%    & 70.5\% (0.53)     &  70.8\% (0.22)        \\ 
    INT8            & 68.4\%    & 70.6\% (0.25)     &  70.6\% (0.07)        \\ 
    \bottomrule
    \end{tabular}
\end{wraptable}

\fakeparagraph{Results}
We compare \sysname with the vanilla-updates, \ie the models are trained from a random initialization with the corresponding methods on all available samples. 
The test accuracy and the ratio of (server-to-edge) communication cost related to full updating on FP32 dense model are reported in \tabref{tab:singleround}. 
Results show that \sysname often yields a higher accuracy than vanilla updating while requiring substantially lower communication cost. 

\section{Conclusion}
\label{sec:conclusion}

In this paper, we present the weight-wise deep partial updating paradigm, motivated by the fact that continuous full weight updating may be impossible in many edge intelligence scenarios. 
We present \sysname through analytically upper-bounding the loss difference between partial updating and full updating. 
\sysname only updates the weights that make the largest contributions to the upper bound, while reuses the other weights that have less impact on the loss reduction. 
Extensive experimental results demonstrate the efficacy of \sysname which achieves a high inference accuracy while updating a rather small number of weights.

\section*{Acknowledgement}
Part of Zhongnan Qu and Lothar Thiele's work was supported by the Swiss National Science Foundation in the context of the NCCR Automation. 
Part of Cong Liu's work was supported by NSF CNS 2135625, CPS 2038727, CNS Career 1750263, and a Darpa Shell grant.

\clearpage
%
%
\bibliographystyle{splncs04}
\bibliography{cites}

\clearpage
\appendix
\appendix

\section{Pseudocodes}
\label{app:pseudocodes}

\subsection{Global Contribution Partial Updating}
\label{app:gcpu}

The magnitude pruning method prunes (\ie set as zero) weights with the lowest magnitudes in a model, which often yields a good trade-off between the model accuracy and the number of zero’s weights \cite{bib:ICLR20:Renda}.
We adapt the magnitude pruning proposed in \cite{bib:ICLR20:Renda} to prune the incremental weights $\delta\bm{w}^{\mathrm{f}}$. 
Specially, the elements with the smallest absolute values in $\delta\bm{w}^{\mathrm{f}}$ are set to zero (also rewinding), while the remaining weights are further sparsely fine-tuned with the same learning rate schedule as training $\bm{w}^{\mathrm{f}}$.

\begin{algorithm}[htbp!]
    \caption{Global Contribution Partial Updating (Prune Incremental Weights)}\label{alg:gcpu}
    \KwIn{weights $\bm{w}$, updating ratio $k$, learning rate $\{\alpha^q\}_{q=1}^{Q}$ in $Q$ iterations} 
    \KwOut{weights $\widetilde{\bm{w}}$}
    \tcc{The first step: full updating and rewinding}
    Initiate $\bm{w}^0=\bm{w}$\;
    \For {$q \leftarrow 1$ \KwTo $Q$} {
        Compute the loss gradient $\bm{g}(\bm{w}^{q-1})=\partial\ell(\bm{w}^{q-1})/\partial\bm{w}^{q-1}$\;
        Compute the optimization step with learning rate $\alpha^q$ as $\Delta\bm{w}^{q}$\;
        Update $\bm{w}^{q} = \bm{w}^{q-1} + \Delta\bm{w}^{q}$\;
    }
    Set $\bm{w}^{\mathrm{f}}=\bm{w}^{Q}$ and get $\delta\bm{w}^{\mathrm{f}} = \bm{w}^{\mathrm{f}}-\bm{w}$\;
    Compute $\bm{c}^{\mathrm{global}}=\delta\bm{w}^{\mathrm{f}}\odot\delta\bm{w}^{\mathrm{f}}$ and sort in descending order\;
    Create a binary mask $\bm{m}$ with $1$ for the Top-$(k \cdot I)$ indices, $0$ for others\;
    \tcc{The second step: sparse fine-tuning}
    Initiate $\widetilde{\delta\bm{w}}=\delta\bm{w}^{\mathrm{f}}\odot\bm{m}$ and $\widetilde{\bm{w}} = \bm{w}+\widetilde{\delta\bm{w}}$\;
    \For {$q \leftarrow 1$ \KwTo $Q$} {
        Compute the optimization step on $\widetilde{\bm{w}}$ with learning rate $\alpha^q$ as $\Delta\widetilde{\bm{w}}^{q}$\; 
        Update $\widetilde{\delta\bm{w}}=\widetilde{\delta\bm{w}}+\Delta\widetilde{\bm{w}}^{q}\odot\bm{m}$ and $\widetilde{\bm{w}}=\bm{w}+\widetilde{\delta\bm{w}}$\;
    }
\end{algorithm}

In comparison to traditional pruning on weights, pruning on incremental weights has a different start point. 
Traditional pruning on weights first trains randomly initialized weights (a zero-initialized model cannot be trained due to the symmetry), and then prunes the weights with the smallest magnitudes.
However, the increment of weights $\delta\bm{w}^\mathrm{f}$ is initialized with zero in Alg. \ref{alg:gcpu}, since the first step starts from $\bm{w}$.
This implies that pruning $\delta\bm{w}^\mathrm{f}$ has the same functionality as rewinding these weights to their initial values in $\bm{w}$.

\section{Complexity Analysis}
\label{app:complexity}

\fakeparagraph{Algorithm \ref{alg:dpu}: Deep Partial Updating}
Recall that the total number of weights vector is denoted as $I$.
In $Q$ optimization iterations during the first step, Alg. \ref{alg:dpu} introduces an extra time complexity of $O(QI)$, and an extra space complexity of $O(I)$ related to the original optimizer.
The rest of the first step takes a time complexity of $O(I\cdot\mathrm{log}(I))$ and a space complexity of $O(I)$ (\eg using heap sort or quick sort).
In $Q$ optimization iterations during the second step, Alg. \ref{alg:dpu} introduces an extra time complexity of $O(QI)$, and an extra space complexity of $O(I)$ related to the original optimizer.
Thus, a total extra time complexity is $O(2QI+I\cdot\mathrm{log}(I))$ and a total extra space complexity is $O(I)$.

\fakeparagraph{Algorithm \ref{alg:gcpu}: Global Contribution Partial Updating}
Recall that the total number of weights vector is denoted as $I$.
In $Q$ optimization iterations during the first step, Alg. \ref{alg:gcpu} does not introduce extra time complexity or extra space complexity related to the original optimizer.
The rest of the first step takes a time complexity of $O(I\cdot\mathrm{log}(I))$ and a space complexity of $O(I)$ (\eg using heap sort or quick sort).
In $Q$ optimization iterations during the second step, Alg. \ref{alg:gcpu} introduces an extra time complexity of $O(QI)$, and an extra space complexity of $O(I)$ related to the original optimizer.
Thus, a total extra time complexity is $O(QI+I\cdot\mathrm{log}(I))$ and a total extra space complexity is $O(I)$.

\section{Implementation Details}
\label{app:implementation}

\subsection{MLP on MNIST}
\label{app:mlp}

The MNIST dataset \cite{bib:MNIST} consists of $28\times28$ gray scale images in 10 digit classes. 
It contains a training dataset with 60000 data samples, and a test dataset with 10000 data samples. 
We use the original training dataset for training; and randomly select 3000 samples in the original test dataset for validation, and the rest 7000 samples for testing. 
We use a mini-batch with size of 128 training on 1 GeForce RTX 3090 GPU.
We use Adam variant of SGD as the optimizer, and use all default parameters provided by Pytorch. 
The number of training epochs is chosen as 60 at each round. 
The initial learning rate is $0.005$, and it decays with a factor of 0.1 every $20$ epochs.
The used MLP contains two hidden layers, and each hidden layer contains 512 hidden units.
The input is a 784-dim tensor consisting of all pixel values in each image.
We use ReLU as the activation function, and use a softmax function as the non-linearity of the last layer (\ie the output layer) in the entire paper.
All weights in MLP need around $2.67$MB.
Each data sample needs $0.784$KB.
The size of MLP equals around 3400 data samples. 
The used MLP architecture is presented as, 2$\times$512FC - 10SVM.

\subsection{VGGNet on CIFAR10}
\label{app:vgg}
The CIFAR10 dataset \cite{bib:cifar10} consists of $32\times32$ color images in 10 object classes. 
It contains a training dataset with 50000 data samples, and a test dataset with 10000 data samples. 
We use the original training dataset for training; and randomly select 3000 samples in the original test dataset for validation, and the rest 7000 samples for testing.
We use a mini-batch with size of 128 training on 1 GeForce RTX 3090 GPU.
We use Adam variant of SGD as the optimizer, and use all default parameters provided by Pytorch. 
The number of training epochs is chosen as 60 at each round. 
The initial learning rate is $0.005$, and it decays with a factor of 0.2 every $20$ epochs.
The used VGGNet is widely adopted in many previous compression works \cite{bib:NIPS15:Courbariaux,bib:ECCV16:Rastegari,bib:CVPR20:Qu}, which is a modified version of the original VGG \cite{bib:ICLR15:Simonyan}.
All weights in VGGNet need around $56.09$MB.
Each data sample needs $3.072$KB.
The size of VGGNet equals around 18200 data samples.
The used VGGNet architecture is presented as, 2$\times$128C3 - MP2 - 2$\times$256C3 - MP2 - 2$\times$512C3 - MP2 - 2$\times$1024FC - 10SVM.

\subsection{ResNet56 on CIFAR100}
\label{app:resnet56}
Similar as CIFAR10, the CIFAR100 dataset \cite{bib:cifar10} consists of $32\times32$ color images in 100 object classes. 
It contains a training dataset with 50000 data samples, and a test dataset with 10000 data samples. 
We use the original training dataset for training; and randomly select 3000 samples in the original test dataset for validation, and the rest 7000 samples for testing.
We use a mini-batch with size of 128 training on 1 GeForce RTX 3090 GPU.
We use Nesterov SGD with weight decay 0.0001 as the optimizer, and use all default parameters provided by Pytorch. 
The number of training epochs is chosen as 100 at each round. 
The initial learning rate is $0.1$, and it decays with the cosine annealing schedule.
The ResNet56 used in our experiments is proposed in \cite{bib:CVPR16:He}.
All weights in ResNet56 need around $3.44$MB.
Each data sample needs $3.072$KB.
The size of ResNet56 equals around 1100 data samples.

\subsection{MobileNetV1 on ImageNet}
\label{app:mobilenetv1}
The ImageNet dataset \cite{bib:ILSVRC15} consists of high-resolution color images in 1000 object classes. 
It contains a training dataset with $1.28$ million data samples, and a validation dataset with 50000 data samples. 
Following the commonly used pre-processing \cite{bib:torchResNet}, each sample (single image) is randomly resized and cropped into a $224\times224$ color image.
We use the original training dataset for training; and randomly select 15000 samples in the original validation dataset for validation, and the rest 35000 samples for testing.
We use a mini-batch with size of 1024 training on 4 GeForce RTX 3090 GPUs. 
We use Nesterov SGD with weight decay 0.0001 as the optimizer, and use all default parameters provided by Pytorch. 
The number of training epochs is chosen as 150 at each round. 
The initial learning rate is $0.5$, and it decays with the cosine annealing schedule.
The MobileNetV1 used in our experiments is proposed in \cite{bib:arXiv17:Howard}.
All weights in MobileNetV1 need around $16.93$MB.
Each data sample needs $150.528$KB.
The size of MobileNetV1 equals around 340 data samples.

\begin{figure}[htbp!]
    \setlength\tabcolsep{\imgtabcolsep}
    \centering
    \begin{tabular}{ccc}
        \textbf{~~~~\{1000,~5000\}}     & \textbf{~~~~\{5000,~1000\}}       & \textbf{~~~~\{1000,~1000\}}       \\
        \tabimgb{figure/1000-5000-f}    & \tabimgb{figure/5000-1000-f}      & \tabimgb{figure/1000-1000-f}      \\
        \tabimgb{figure/1000-5000-fv}   & \tabimgb{figure/5000-1000-fv}     & \tabimgb{figure/1000-1000-fv}     \\
    \end{tabular}
    \caption{Comparing full updating methods with different initialization at each round.}
    \label{fig:fullupdating}
\end{figure}

\section{Full Updating}
\label{app:fullupdating}

\fakeparagraph{Settings}
In this experiment, we compare full updating with different initialization at each round to confirm the best-performed full updating baseline. 
The compared full updating methods include, (\textit{i}) the model is trained from different random initialization at each round; (\textit{ii}) the model is trained from a same random initialization at each round, \ie with the same random seed; (\textit{iii}) the model is trained from the weights $\bm{w}^{r-1}$ of the last round at each round.
The experiments are conducted on VGGNet using CIFAR10 dataset with different amounts of training samples $\{|\mathcal{D}^1|,|\delta\mathcal{D}^r|\}$.
Each experiment runs for three times using random data samples.

\fakeparagraph{Results}
We report the mean and the standard deviation of test accuracy (over three runs) under different initialization in \figref{fig:fullupdating}.
The results show that training from a same random initialization yields a similar accuracy level while sometimes also a lower variance, as training from different random initialization at each round.
In comparison to training from scratch (\ie random initialization), training from $\bm{w}^{r-1}$ may yield a higher accuracy in the first few rounds; yet training from scratch can always outperform after a large number of rounds.
Thus, in this paper, we adopt training from a same random initialization at each round, \ie (\textit{ii}), as the baseline of full updating.

\section{Number of Rounds for Re-Initialization}
\label{app:reinit}

\fakeparagraph{Settings}
In these experiments, we re-initialize the model every $n$ rounds under different partial updating settings to determine a heuristic rule to set the number of rounds for re-initialization.
We conduct experiments on VGGNet using CIFAR10 dataset, with different amounts of training samples $\{|\mathcal{D}^1|,|\delta\mathcal{D}^r|\}$ and different updating ratios $k$.
Every $n$ rounds, the model is (re-)initialized again from a same random model (as mentioned in \ref{app:fullupdating}), then partially updated in the next $n$ rounds with Alg. \ref{alg:dpu}.
We choose $n=1,5,10,20$.
Specially, $n=1$ means that the model is partially updated from the same random model every round, \ie without reusing the learned knowledge at all.
Each experiment runs three times using random data samples.

\begin{figure}[tbp!]
    \setlength\tabcolsep{\imgtabcolsep}
    \centering
    \begin{tabular}{m{0.3cm}ccc}
                                          & \textbf{~~~~\{1000,~5000\}}    & \textbf{~~~~\{5000,~1000\}}    & \textbf{~~~~\{1000,~1000\}}         \\ 
        \rotatebox{90}{\textbf{0.01}}     & \tabimga{figure/1000-5000-1-r} & \tabimga{figure/5000-1000-1-r} & \tabimga{figure/1000-1000-1-r}      \\ 
        \rotatebox{90}{\textbf{0.05}}     & \tabimga{figure/1000-5000-2-r} & \tabimga{figure/5000-1000-2-r} & \tabimga{figure/1000-1000-2-r}      \\
        \rotatebox{90}{\textbf{0.1}}      & \tabimga{figure/1000-5000-3-r} & \tabimga{figure/5000-1000-3-r} & \tabimga{figure/1000-1000-3-r}      \\ 
    \end{tabular}
    \caption{Comparison w.r.t. the mean accuracy when \sysname is re-initialized every $n$ rounds ($n=1,5,10,20$) under different $\{|\mathcal{D}^1|,|\delta\mathcal{D}^r|\}$ and updating ratio ($k=0.01,0.05,0.1$) settings.}
    \label{fig:partial_reinit}
\end{figure}

\begin{figure}[htbp!]
    \setlength\tabcolsep{\imgtabcolsep}
    \centering
    \begin{tabular}{m{0.3cm}ccc}
                                          & \textbf{~~~~\{1000,~5000\}}     & \textbf{~~~~\{5000,~1000\}}     & \textbf{~~~~\{1000,~1000\}}       \\ 
        \rotatebox{90}{\textbf{0.01}}     & \tabimga{figure/1000-5000-1-rp} & \tabimga{figure/5000-1000-1-rp} & \tabimga{figure/1000-1000-1-rp}   \\ 
        \rotatebox{90}{\textbf{0.05}}     & \tabimga{figure/1000-5000-2-rp} & \tabimga{figure/5000-1000-2-rp} & \tabimga{figure/1000-1000-2-rp}   \\
        \rotatebox{90}{\textbf{0.1}}      & \tabimga{figure/1000-5000-3-rp} & \tabimga{figure/5000-1000-3-rp} & \tabimga{figure/1000-1000-3-rp}   \\
        \rotatebox{90}{\textbf{1}}        & \tabimga{figure/1000-5000-fr}   & \tabimga{figure/5000-1000-fr}   & \tabimga{figure/1000-1000-fr}     \\
    \end{tabular}
    \caption{Comparison w.r.t. the mean accuracy when \sysname is re-initialized every $n$ rounds ($n=1,5,10,20$) under different $\{|\mathcal{D}^1|,|\delta\mathcal{D}^r|\}$ and updating ratio ($k=0.01,0.05,0.1$ and full updating $k=1$) settings.}
    \label{fig:partial_reinit_savecomm}
\end{figure}

\fakeparagraph{Results}
We plot the mean test accuracy along rounds in \figref{fig:partial_reinit}.
By comparing $n=1$ with other settings, we can conclude that within a certain number of rounds, the current deployed model $\bm{w}^{r-1}$ (\ie the model from the last round) is a better starting point for Alg. \ref{alg:dpu} than a randomly initialized model. 
In other word, partially updating from the last round may yield a higher accuracy than partially updating from a random model with the same training effort.
This is straightforward, since such a model is already pretrained on a subset of the currently available data samples, and the previous learned knowledge could help the new training process. 
Since all newly collected samples are continuously stored in the server, complete information about the past data samples is available. 
This also makes our setting different from continual learning setting, which aims at avoiding catastrophic forgetting without accessing (at least not all) old data.

Each time the model is re-initialized, the new partially updated model might suffer from an accuracy drop in a few rounds.
Although this accuracy drop may be relieved if we carefully tune the partial updating training scheme every time, this is not feasible regarding a large number of updating rounds.
However, we can simply avoid such an accuracy drop by not updating the model if the validation accuracy does not increase compared to the last round (as discussed in \secref{sec:experiment}).
Note that in this situation, the partially updated weights (as well as the random seed for re-initialization) still need to be sent to the edge devices, since this is an on-going training process. 
After implementing the above strategy, we plot the mean accuracy in \figref{fig:partial_reinit_savecomm}. 
In addition, we also add the related results of full updating in \figref{fig:partial_reinit_savecomm}, where the model is fully updated and is re-initialized every $n$ rounds from a same random model.  
Note that full updating with re-initialization every round ($n=1$) is exactly the same as ``same rand init.'' in \figref{fig:fullupdating} in \ref{app:fullupdating}.
From \figref{fig:partial_reinit_savecomm}, we can conclude that the model needs to be re-initialized more frequently in the first several rounds than in the following rounds to achieve a higher accuracy level.
The model also needs to be re-initialized more frequently with a large partial updating ratio $k$.
Particularly, the ratio between the number of current data samples and the number of following collected data samples has a larger impact than the updating ratio.

Thus, we propose to re-initialize the model as long as the number of total newly collected data samples exceeds the number of samples when the model is re-initialized last time.
For example, assume that at round $r$ the model is randomly (re-)initialized and partially updated from the random model on dataset $\mathcal{D}^r$. 
Then, the model will be re-initialized at round $r+n$, if $|\mathcal{D}^{r+n}|>2\cdot|\mathcal{D}^r|$. 

\section{Additional Multi-Round Updating Results}
\label{app:multiround}

\subsection{Experiments on Total Communication Cost Reduction}
\label{app:totalcost}

\fakeparagraph{Settings}
In this experiment, we show the advantages of \sysname in terms of the total communication cost reduction, as \sysname has no impact on the edge-to-server communication which may involve sending new data samples collected on edge nodes.
The total communication cost includes both the edge-to-server communication and the server-to-edge communication.
Here we assume that all samples in $\delta\mathcal{D}^r$ are collected by $N$ edge nodes during all rounds and sent to the server on a per-round basis. 
In other words, the first stage (see in \secref{sec:introduction}) is anyway necessary for sending new training data to the server.
For clarity, let $S_d$ denote the data size of each training sample. 
During round $r$, we define the per-node total communication cost under \sysname as $S_d\cdot|\delta\mathcal{D}^r|/N + (S_w \cdot k \cdot I + S_x(k) \cdot I)$.
Similarly, the per-node total communication cost under full updating is defined as $S_d\cdot|\delta\mathcal{D}^r|/N + S_w \cdot I$.

In order to simplify the demonstration, we consider the scenario where $N$ nodes send a certain amount of data samples to the server in $R-1$ rounds, namely $\sum_{r=2}^R |\delta \mathcal{D}^r|$ (see \secref{sec:initialization}).
Thus, the average data size transmitted from each node to the server in all rounds is $\sum_{r=2}^R S_d\cdot|\delta \mathcal{D}^r|/N$. 
A larger $N$ implies a fewer amount of transmitted data from each node to the server.

\begin{figure}[tbp!]
    \setlength\tabcolsep{\imgtabcolsep}
    \centering
    \begin{tabular}{ccc}
        \tabimgb{figure/mlp-N}    & \tabimgb{figure/vgg-N}      & \tabimgb{figure/resnet56-N}      \\ 
    \end{tabular}
    \caption{The ratio, between the total communication cost (over all rounds) under \sysname and that under full updating, varies with the number of edge nodes $N$.}
    \label{fig:communicationratio}
\end{figure}

\fakeparagraph{Results}
We report the ratio of the total communication cost over all rounds required by \sysname related to full updating, when \sysname achieves a similar accuracy level as full updating (corresponding to three evaluations in \figref{fig:multiround}).
The ratio clearly depends on $\sum_{r=2}^R S_d\cdot|\delta \mathcal{D}^r|/N$, \ie the number of nodes $N$.
The relation between the ratio and $N$ is plotted in \figref{fig:communicationratio}. 

\sysname can reduce up to $88.2\%$ of the total communication cost even for a single node.
Single node corresponds to the largest data size during edge-to-serve transmission per node, \ie the worst case.  
Moreover, \sysname tends to be more beneficial when the size of data transmitted by each node to the server becomes smaller. 
This is intuitive because in this case the server-to-edge communication (thus the reduction due to \sysname) dominants in the entire communication.

\subsection{Impact due to Varying Number of Data Samples and Updating Ratios}
\label{app:ablation}

\fakeparagraph{Settings}
In this section, we show that \sysname outperforms other baselines under varying number of training samples and updating ratios in multi-round updating.
We also conduct ablations concerning the re-initialization of weights discussed in \secref{sec:initialization}.
We implement \sysname with and without re-initialization, GCPU with and without re-initialization, RPU, and Pruning \cite{bib:ICLR20:Renda} (see more details in \secref{sec:multiround}) on VGGNet using CIFAR10 dataset. 
We compare these methods with different amounts of samples $\{|\mathcal{D}^1|,|\delta\mathcal{D}^r|\}$ and different updating ratios $k$.
Each experiment runs three times using random data samples.

\begin{figure}[tbp!]
    \setlength\tabcolsep{\imgtabcolsep}
    \centering
    \begin{tabular}{m{0.3cm}ccc}
                                                  & \textbf{~~~~~~0.01}                   & \textbf{~~~~~~0.05}                   & \textbf{~~~~~~0.1}                        \\ 
        \rotatebox{90}{\textbf{\{1000,1000\}}}    & \tabimga{figure/1000-1000-1-pruningd} & \tabimga{figure/1000-1000-2-pruningd} & \tabimga{figure/1000-1000-3-pruningd}     \\ 
        \rotatebox{90}{\textbf{\{1000,5000\}}}    & \tabimga{figure/1000-5000-1-pruningd} & \tabimga{figure/1000-5000-2-pruningd} & \tabimga{figure/1000-5000-3-pruningd}     \\ 
        \rotatebox{90}{\textbf{\{5000,1000\}}}    & \tabimga{figure/5000-1000-1-pruningd} & \tabimga{figure/5000-1000-2-pruningd} & \tabimga{figure/5000-1000-3-pruningd}     \\
        \rotatebox{90}{\textbf{\{1000,500\}}}     & \tabimga{figure/1000-500-1-pruningd}  & \tabimga{figure/1000-500-2-pruningd}  & \tabimga{figure/1000-500-3-pruningd}      \\
        \rotatebox{90}{\textbf{\{500,1000\}}}     & \tabimga{figure/500-1000-1-pruningd}  & \tabimga{figure/500-1000-2-pruningd}  & \tabimga{figure/500-1000-3-pruningd}      \\
    \end{tabular}
    \caption{Comparison w.r.t. the mean accuracy difference (full updating as the reference) under different $\{|\mathcal{D}^1|,|\delta\mathcal{D}^r|\}$ and updating ratio ($k=0.01,0.05,0.1$) settings.}
    \label{fig:number_ratio_full_d}
\end{figure}

\begin{figure}[tbp!]
    \setlength\tabcolsep{\imgtabcolsep}
    \centering
    \begin{tabular}{m{0.3cm}ccc}
                                                  & \textbf{~~~~~~0.01}                   & \textbf{~~~~~~0.05}                   & \textbf{~~~~~~0.1}                       \\ 
        \rotatebox{90}{\textbf{\{1000,1000\}}}    & \tabimga{figure/1000-1000-1-pruning}  & \tabimga{figure/1000-1000-2-pruning}  & \tabimga{figure/1000-1000-3-pruning}     \\ 
        \rotatebox{90}{\textbf{\{1000,5000\}}}    & \tabimga{figure/1000-5000-1-pruning}  & \tabimga{figure/1000-5000-2-pruning}  & \tabimga{figure/1000-5000-3-pruning}     \\ 
        \rotatebox{90}{\textbf{\{5000,1000\}}}    & \tabimga{figure/5000-1000-1-pruning}  & \tabimga{figure/5000-1000-2-pruning}  & \tabimga{figure/5000-1000-3-pruning}     \\
        \rotatebox{90}{\textbf{\{1000,500\}}}     & \tabimga{figure/1000-500-1-pruning}   & \tabimga{figure/1000-500-2-pruning}   & \tabimga{figure/1000-500-3-pruning}      \\
        \rotatebox{90}{\textbf{\{500,1000\}}}     & \tabimga{figure/500-1000-1-pruning}   & \tabimga{figure/500-1000-2-pruning}   & \tabimga{figure/500-1000-3-pruning}      \\
    \end{tabular}
    \caption{Comparison w.r.t. the mean accuracy under different $\{|\mathcal{D}^1|,|\delta\mathcal{D}^r|\}$ and updating ratio ($k=0.01,0.05,0.1$) settings.}
    \label{fig:number_ratio_full}
\end{figure}

\begin{figure}[tbp!]
    \setlength\tabcolsep{\imgtabcolsep}
    \centering
    \begin{tabular}{m{0.3cm}ccc}
                                                  & \textbf{~~~~~~0.01}                    & \textbf{~~~~~~0.05}                    & \textbf{~~~~~~0.1}                        \\ 
        \rotatebox{90}{\textbf{\{1000,1000\}}}    & \tabimga{figure/1000-1000-1-pruningv}  & \tabimga{figure/1000-1000-2-pruningv}  & \tabimga{figure/1000-1000-3-pruningv}     \\ 
        \rotatebox{90}{\textbf{\{1000,5000\}}}    & \tabimga{figure/1000-5000-1-pruningv}  & \tabimga{figure/1000-5000-2-pruningv}  & \tabimga{figure/1000-5000-3-pruningv}     \\ 
        \rotatebox{90}{\textbf{\{5000,1000\}}}    & \tabimga{figure/5000-1000-1-pruningv}  & \tabimga{figure/5000-1000-2-pruningv}  & \tabimga{figure/5000-1000-3-pruningv}     \\
        \rotatebox{90}{\textbf{\{1000,500\}}}     & \tabimga{figure/1000-500-1-pruningv}   & \tabimga{figure/1000-500-2-pruningv}   & \tabimga{figure/1000-500-3-pruningv}      \\
        \rotatebox{90}{\textbf{\{500,1000\}}}     & \tabimga{figure/500-1000-1-pruningv}   & \tabimga{figure/500-1000-2-pruningv}   & \tabimga{figure/500-1000-3-pruningv}      \\
    \end{tabular}
    \caption{Comparison w.r.t. the standard deviation of accuracy under different $\{|\mathcal{D}^1|,|\delta\mathcal{D}^r|\}$ and updating ratio ($k=0.01,0.05,0.1$) settings.}
    \label{fig:number_ratio_full_v}
\end{figure}

\fakeparagraph{Results}
We compare the difference between the accuracy under each partial updating method and that under full updating. 
The mean accuracy difference (over three runs) is plotted in \figref{fig:number_ratio_full_d}.
As seen in \figref{fig:number_ratio_full_d}, \sysname (with re-initialization) always achieves the highest accuracy.
\sysname also significantly outperforms the pruning method, especially under a small updating ratio. 
Note that we preferred a smaller updating ratio in our context because it explores the limits of the approach and it indicates that we can improve the deployed model more frequently with the same accumulated server-to-edge communication cost.
In addition, we also plot the mean and standard deviation of the absolute accuracy of these methods (including full updating) in \figref{fig:number_ratio_full} and \figref{fig:number_ratio_full_v}, respectively.
The dashed curves and the solid curves with the same color in these figures can be viewed as the ablation study of our re-initialization scheme. 
Particularly given a large number of rounds, it is critical to re-initialize the start point $\bm{w}^{r-1}$ after several rounds (as discussed in \secref{sec:initialization}). 

\section{Impacts from Global/Local Contributions}
\label{app:impact}

\subsection{Ablation Studies of Rewinding Metrics}
\label{sec:lossIncr}

\fakeparagraph{Settings} 
We conduct a set of ablation experiments regarding different rewinding metrics discussed in \secref{sec:metric}. 
We compare the influence of the local and global contributions as well as their combination, in terms of the training loss after the rewinding and the final test accuracy. 
We conduct single-round updating on VGGNet.
The initial model are fully trained on a randomly selected dataset of $10^3$ samples. 
After adding $10^3$ new randomly selected samples, we conduct the first step of our approach (see Alg. \ref{alg:dpu}) with all three rewinding metrics, \ie the global contribution, the local contribution, and the combined contribution.
Accordingly, the second step (sparse fine-tuning) is executed.
The experiment is executed over five runs with different random seeds.

\fakeparagraph{Results}
The training loss after rewinding (\ie $\ell(\bm{w}+\delta\bm{w}^\mathrm{f}\odot\bm{m})$) and the final test accuracy after sparse fine-tuning (\ie at $\widetilde{\bm{w}}$) are reported in \tabref{tab:lossincr}.
We use the form of mean $\pm$ standard deviation.
As seen in the table, the combined contribution always yields a lower or similar training loss after rewinding compared to the other two metrics.
The smaller deviation also indicates that adopting the combined contribution yields more robust results.
This demonstrates the effectiveness of our proposed metric, \ie the combined contribution to the analytical upper bound on loss reduction.
Rewinding with the combined contribution also acquires a higher final accuracy, which in turn verifies the hypothesis we made for partial updating, a weight shall be updated only if it has a large contribution to the loss reduction.

\begin{table}[htbp!]
    \centering
    \caption{Comparing training loss after rewinding and the final test accuracy under different metrics.} 
    \label{tab:lossincr}
    \begin{tabular}{cccc}
        \toprule
        \multirow{2}{*}{$k$}  & \multicolumn{3}{c}{Training loss at $\bm{w}+\delta\bm{w}^\mathrm{f}\odot\bm{m}$ ~~~~(Test accuracy at $\widetilde{\bm{w}}$)}           \\ \cline{2-4} 
                              & \multicolumn{1}{c}{Global}                   & \multicolumn{1}{c}{Local}                & \multicolumn{1}{c}{Combined}                 \\ \hline
        0.01                  & $3.04\pm0.07$ $(55.0\pm0.1\%)$               & $\bm{2.59}\pm0.08$ $(55.6\pm0.1\%)$      & $2.66\pm0.09$ $(\bm{56.5}\pm0.0\%)$          \\
        0.05                  & $2.51\pm0.06$ $(57.3\pm0.2\%)$               & $1.80\pm0.10$ $(57.8\pm0.1\%)$           & $\bm{1.67}\pm0.06$ $(\bm{58.2}\pm0.1\%)$     \\
        0.1                   & $2.03\pm0.05$ $(58.3\pm0.0\%)$               & $1.34\pm0.08$ $(59.0\pm0.1\%)$           & $\bm{0.99}\pm0.03$ $(\bm{59.0}\pm0.1\%)$     \\
        0.2                   & $1.20\pm0.05$ $(59.0\pm0.1\%)$               & $0.74\pm0.03$ $(59.6\pm0.2\%)$           & $\bm{0.42}\pm0.01$ $(\bm{60.1}\pm0.2\%)$     \\ 
        \bottomrule
    \end{tabular}
\end{table}

\subsection{Balancing between Global and Local Contributions}
\label{app:balancing}

\fakeparagraph{Settings} 
In \equref{eq:combined}, the combined contribution is calculated by adding both normalized contributions together.
However, both normalized contributions may have different importance when determining the critical weights.
In order to investigate which one plays a more essential role in the combined contribution, we introduce another hyper-parameter $\lambda$ to tune the proportion of both normalized contributions as
\begin{equation}
    \bm{c}_\lambda = \lambda \cdot \frac{1}{\bm{1}^\mathrm{T} \cdot \bm{c}^\mathrm{global}} \bm{c}^\mathrm{global} + (1-\lambda) \cdot \frac{1}{\bm{1}^\mathrm{T} \cdot \bm{c}^\mathrm{local}} \bm{c}^\mathrm{local}
\end{equation}
Note that the combined contribution $\bm{c}$ used in the previous experiments is the same as $\bm{c}_\lambda$ when $\lambda=0.5$, since only the order matters when determining the critical weights.
We implement partial updating methods with the rewinding metric $\bm{c}_\lambda$ under different values of $\lambda$. 
We compare these methods under updating ratios $k=0.01,0.05,0.1$ and different $\{|\mathcal{D}^1|,|\delta\mathcal{D}^r|\}$ settings on VGGNet using CIFAR10 dataset, and with the re-initialization scheme described in \secref{sec:initialization}.
Each experiment runs three times using random data samples.

\fakeparagraph{Results}
To clearly illustrate the impact of $\lambda$, we compare the difference between the accuracy under partial updating methods with various $\lambda$ and that under full updating. The mean accuracy difference (over three runs) are plotted in \figref{fig:combinedratio}.
As seen in \figref{fig:combinedratio}, $\lambda=0.5$ always obtains the best performance in general, especially when the updating ratio is small.
Thus, in the following experiments, we fix this hyper-parameter $\lambda$ as 0.5. 
In other words, the combined contribution is chosen as 
\begin{equation}
    \bm{c}_\lambda(\lambda=0.5) = 0.5 \cdot \frac{1}{\bm{1}^\mathrm{T} \cdot \bm{c}^\mathrm{global}} \bm{c}^\mathrm{global} + 0.5 \cdot \frac{1}{\bm{1}^\mathrm{T} \cdot \bm{c}^\mathrm{local}} \bm{c}^\mathrm{local}
\end{equation}
which has exactly the same functionality as \equref{eq:combined}. 
Note that it may be possible to manually find another hyper-parameter $\lambda$ that achieves better performance in certain cases. 
However, setting $\lambda$ as 0.5 already yields a satisfactory performance, and can avoid meticulous and computationally expensive hyper-parameter tuning in a large number of updating rounds.

\begin{figure}[htbp!]
    \setlength\tabcolsep{\imgtabcolsep}
    \centering
    \begin{tabular}{m{0.3cm}ccc}
                                                  & \textbf{~~~~~~0.01}                 & \textbf{~~~~~~0.05}                 & \textbf{~~~~~~0.1}                      \\ 
        \rotatebox{90}{\textbf{\{1000,1000\}}}    & \tabimga{figure/1000-1000-1-ratiod} & \tabimga{figure/1000-1000-2-ratiod} & \tabimga{figure/1000-1000-3-ratiod}     \\ 
        \rotatebox{90}{\textbf{\{1000,5000\}}}    & \tabimga{figure/1000-5000-1-ratiod} & \tabimga{figure/1000-5000-2-ratiod} & \tabimga{figure/1000-5000-3-ratiod}     \\ 
        \rotatebox{90}{\textbf{\{5000,1000\}}}    & \tabimga{figure/5000-1000-1-ratiod} & \tabimga{figure/5000-1000-2-ratiod} & \tabimga{figure/5000-1000-3-ratiod}     \\
    \end{tabular}
    \caption{Comparison w.r.t. the mean accuracy difference (full updating as the reference) under $\lambda=0.5,0.1,0.3,0.7,0.9$. The chosen settings are updating ratios $k=0.01,0.05,0.1$, $\{|\mathcal{D}^1|,|\delta\mathcal{D}^r|\}=\{1000,1000\}, \{1000,5000\}, \{5000,1000\}$.}
    \label{fig:combinedratio}
\end{figure}

\subsection{Number of Updated Weights across Layers under Different Rewinding Metrics}
\label{app:num_weights}
\fakeparagraph{Settings} 
To further study the impact of adopting different rewinding metrics, we show the distribution of updated weights across layers in this section.
We implement partial updating methods with three rewinding metrics (\ie the global contribution, the local contribution, and the combined contribution, see in \secref{sec:metric}) on VGGNet using CIFAR10 dataset. 
We compare these methods with different updating ratios $k$ under $\{|\mathcal{D}^1|,|\delta\mathcal{D}^r|\}=\{1000,1000\}$.
All methods start from the same randomly initialized model, and are re-initialized with this random model according to the proposed scheme in \secref{sec:initialization}.
To study the distribution of updated weights along all rounds, we let the model partially updated in every round even if the model accuracy may degrade for a few rounds due to the re-initialization.

\begin{figure}[!htbp]
    \centering
	\includegraphics[width=0.9\textwidth]{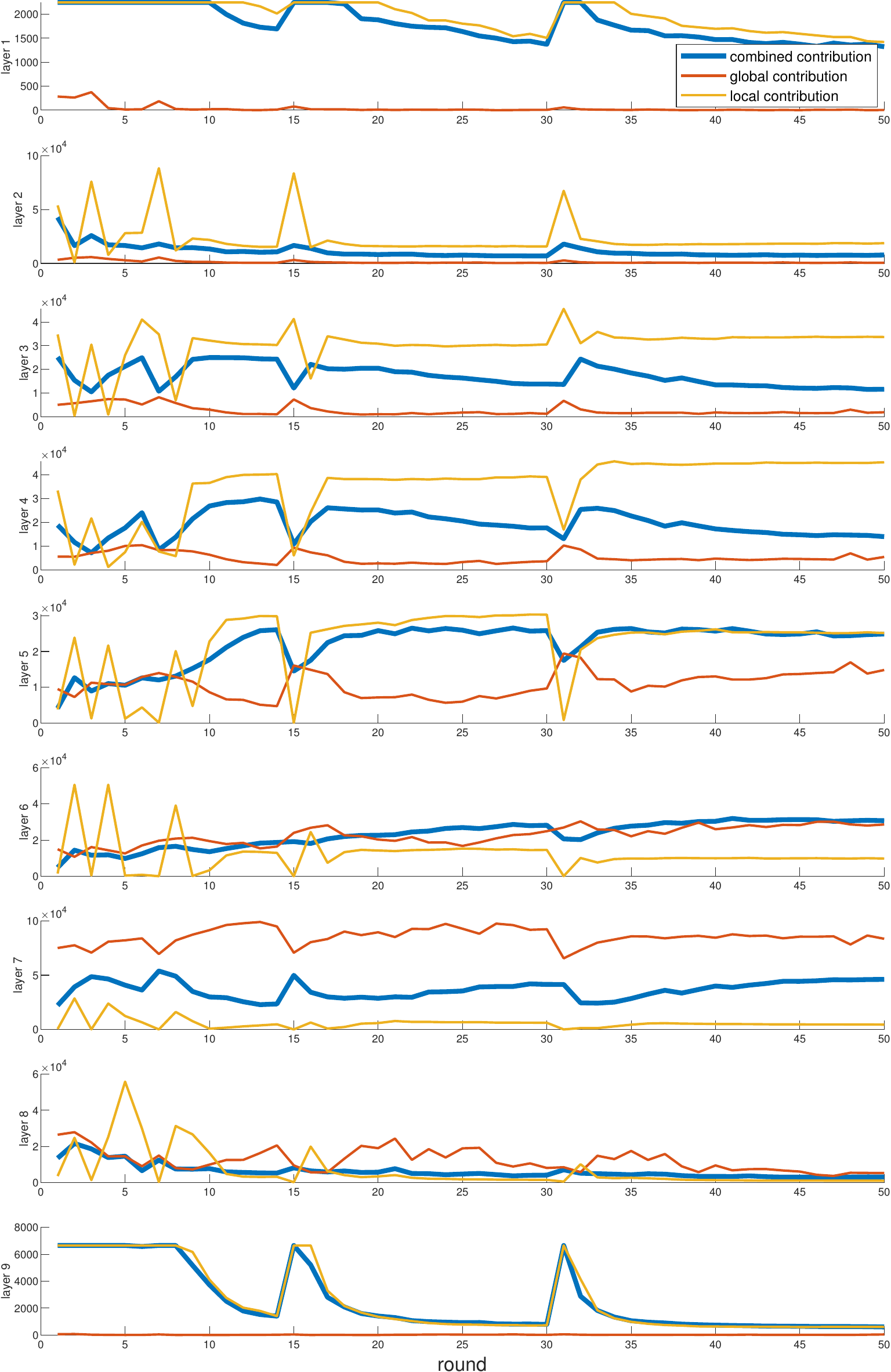}
	\caption{Number of updated weights across all layers (VGGNet) when adopting different rewinding metrics (updating ratio $k=0.01$).}
    \label{fig:updatednum1}
\end{figure}

\begin{figure}[!htbp]
    \centering
	\includegraphics[width=0.9\textwidth]{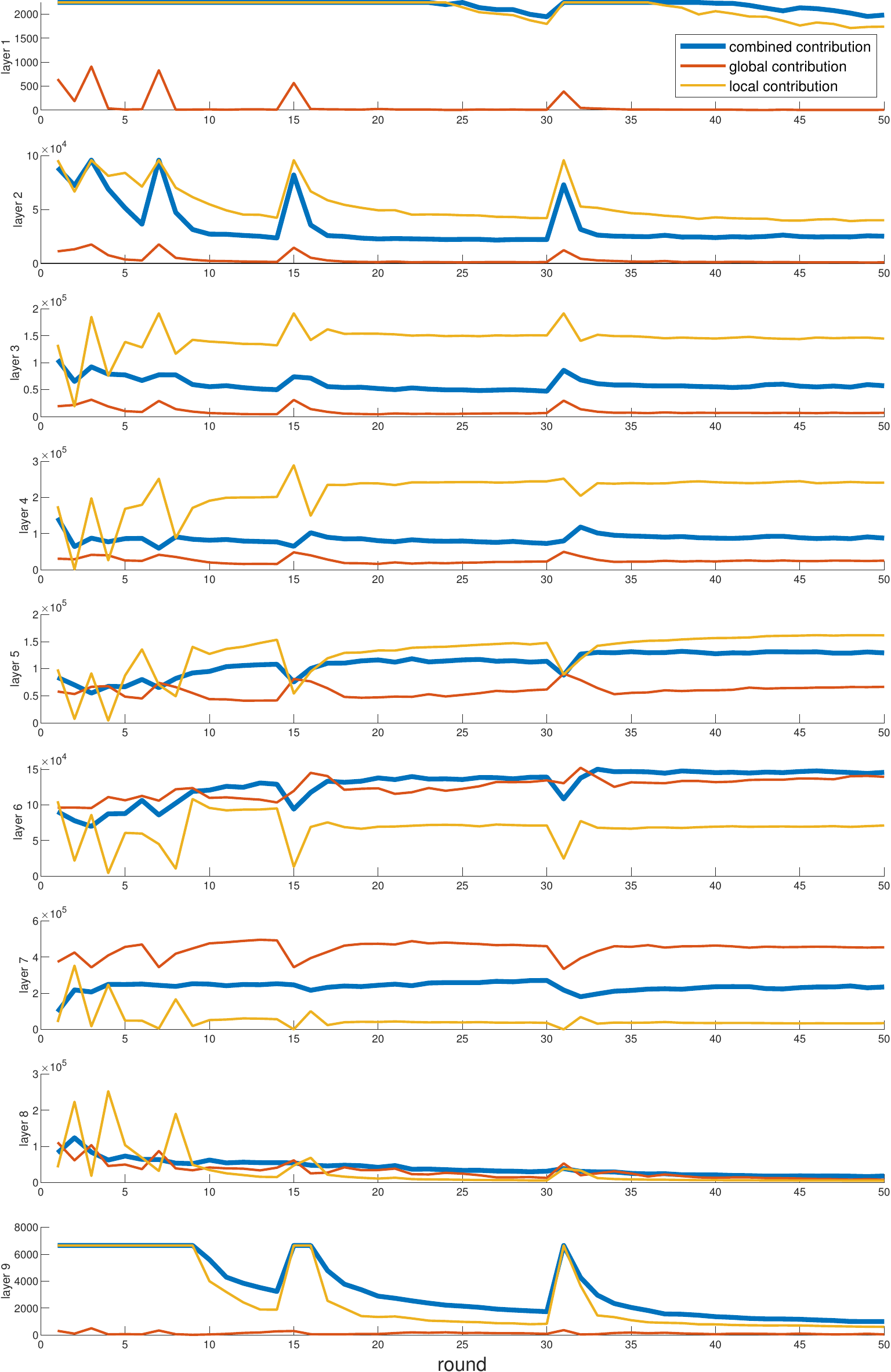}
	\caption{Number of updated weights across all layers (VGGNet) when adopting different rewinding metrics (updating ratio $k=0.05$).}
    \label{fig:updatednum2}
\end{figure}

\begin{figure}[!htbp]
    \centering
	\includegraphics[width=0.9\textwidth]{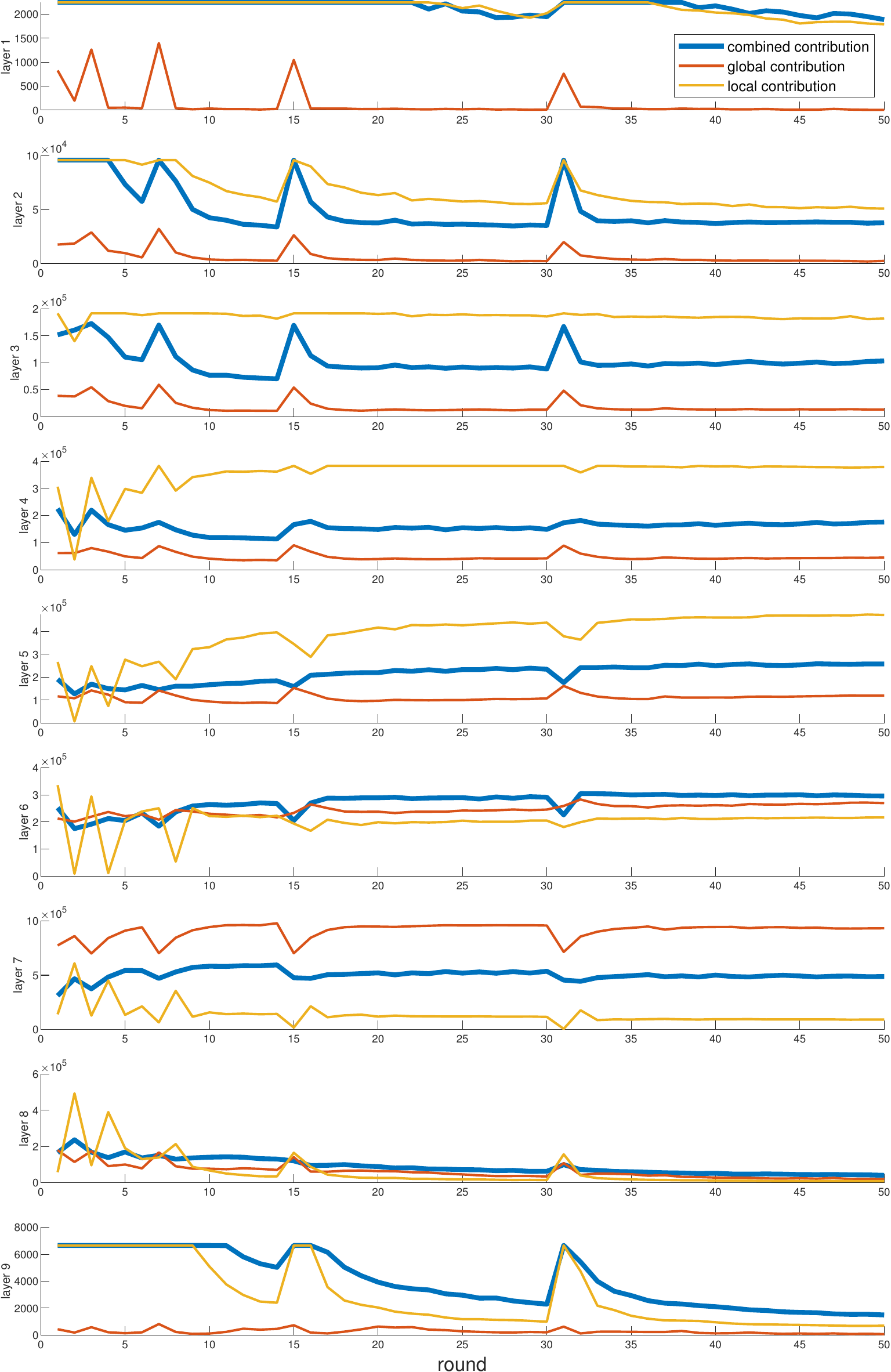}
	\caption{Number of updated weights across all layers (VGGNet) when adopting different rewinding metrics (updating ratio $k=0.1$).}
    \label{fig:updatednum3}
\end{figure}

\fakeparagraph{Results}
We plot the number of updated weights across all layers along rounds, under updating ratio $k=0.01,0.05,0.1$ in \figref{fig:updatednum1}, \figref{fig:updatednum2}, and \figref{fig:updatednum3}, respectively. 
We also plot the corresponding test accuracy along rounds in \figref{fig:updatedacc}.
Generally, the metric of local contribution updates more weights in the first several layers and the last layer while with a large variance along rounds.
On the contrary, global contribution selects more weights in the last several layers (until the penultimate layer) to update.
Combined contribution (the sum of normalized local/global contribution) achieves a more robust and balanced distribution of updated weights across layers than other contributions. 
It also results in the highest accuracy level especially under a small updating ratio.
Intuitively, local contribution can better identify critical weights w.r.t. the loss during training, while global contribution may be more robust for a highly non-convex loss landscape. 
Both metrics may be necessary when selecting weights to rewind.
Note that the proposed combined contribution is not the simple averaging of both local and global contribution.
For example, in ``layer 6'' of \figref{fig:updatednum3}, the number of updated weights by combined contribution already exceeds the other two metrics.

\begin{figure}[htbp]
    \setlength\tabcolsep{\imgtabcolsep}
    \centering
    \begin{tabular}{ccc}
        \textbf{~~~~~~0.01}                         & \textbf{~~~~~~0.05}                           & \textbf{~~~~~~0.1}                            \\
        \tabimgb{figure/1000-1000-1-updated-acc}    & \tabimgb{figure/1000-1000-2-updated-acc}      & \tabimgb{figure/1000-1000-3-updated-acc}      \\
    \end{tabular}
    \caption{The test accuracy of partial updating methods with different rewinding metrics (updating ratio $k=0.01,0.05,0.1$).}
    \label{fig:updatedacc}
\end{figure}

\section{Quantizing and Encoding}
\label{app:pipeline}

\fakeparagraph{Settings}
The updates could also be compressed through quantization and/or encoding to reduce the communication cost.
In this set of experiments, we show these compression techniques (\ie quantization and encoding) are orthogonal to our \sysname.
\cite{bib:ICLR16:Han} proposed that certain types of quantization and encoding could be applied in addition to pruning without hurting the accuracy. 
Following the compression pipeline in \cite{bib:ICLR16:Han}, the resulted sparse updating from our \sysname could also be further quantized and Huffman-encoded.
The overall compression pipeline \textit{in each round} is summarized as follows, (\textit{i}) a partial updating (also sparse updating) is generated from our \sysname; (\textit{ii}) these updates are quantized into 8-bit for each layer, \ie each layer's non-zero values share 256 centroids; (\textit{iii}) the quantized updates are Huffman-encoded; (\textit{iv}) the server sends the encoded updates, the code books (for Huffman-encoding and quantization), as well as the indices to edge devices.

We implement \sysname (with re-initialization see in \secref{sec:initialization}), DPU+Q+E, pruning (a state-of-the-art pruning method proposed in \cite{bib:ICLR20:Renda} see in \secref{sec:samplesRatio}), and pruning+Q+E, to verify that applying Q+E in addition does not bring extra accuracy loss. 
Here, Q stands for the quantization step in (\textit{ii}), and E stands for the encoding step in (\textit{iii}).
We test on VGGNet using CIFAR10 dataset under $\{|\mathcal{D}^1|,|\delta\mathcal{D}^r|\}=\{1000,1000\}$, $\{1000,5000\}$, and $\{5000,1000\}$.
Note that the updating ratio $k$ is set to 0.01, also the most critical case.
Each experiment runs three times using random data samples.

\fakeparagraph{Results}
We plot the mean and standard deviation of test accuracy (over three runs) of these methods in \figref{fig:deepcompression}.
In addition, we also add the baseline of full updating (FU) in the figures for comparison. 
The dashed curves and the solid curves with the same color can be viewed as the ablations of with/without quantization and Huffman-encoding, respectively.
The results reveal that applying these quantization and encoding techniques does not bring performance degradation for both pruning methods and our deep partial updating schemes.
Therefore, the size of transmitted data could be further reduced by quantizing and/or encoding the partial updates resulted from \sysname.  
We report the ratio between the data size of server-to-edge transmission under these above methods and that under full updating in \tabref{tab:deepcompression}. 
Note that the reported ratios are the mean values averaged over all settings in \figref{fig:deepcompression}.
Particularly, in comparison to full updating, our DPU+Q+E can reduce the size of transmitted data by 145$\times$, \ie $99.31\%$, while achieving a similar accuracy level.

\begin{table}[!htb]
    \centering
    \caption{The ratio of communication cost (server-to-edge) over all rounds related to full updating.} 
    \label{tab:deepcompression}
    \begin{tabular}{ccccc}
        \toprule
        Method                    & DPU       & DPU+Q+E   & Pruning   & Pruning+Q+E   \\  \hline
        Ratio                     & 0.0177    & 0.0069    & 0.0174    & 0.0068        \\
    \bottomrule
    \end{tabular}
\end{table}

\begin{figure}[tb]
    \setlength\tabcolsep{\imgtabcolsep}
    \centering
    \begin{tabular}{ccc}
        \textbf{~~~~~\{1000,~5000\}}        & \textbf{~~~~~\{5000,~1000\}}          & \textbf{~~~~~\{1000,~1000\}}      \\
        \tabimgb{figure/1000-5000-1-dc}     & \tabimgb{figure/5000-1000-1-dc}       & \tabimgb{figure/1000-1000-1-dc}   \\
        \tabimgb{figure/1000-5000-1-dcv}    & \tabimgb{figure/5000-1000-1-dcv}      & \tabimgb{figure/1000-1000-1-dcv}  \\
    \end{tabular}
    \caption{Verifying the orthogonality between \sysname (and pruning) and other compression techniques, namely quantization (Q) and Huffman-encoding (E). The chosen settings are updating ratio $k=0.01$, $\{|\mathcal{D}^1|,|\delta\mathcal{D}^r|\}=\{1000,1000\}, \{1000,5000\}, \{5000,1000\}$.}
    \label{fig:deepcompression}
\end{figure}


\end{document}